\documentclass[11pt]{article}
\pdfoutput=1
\usepackage{booktabs}
\usepackage{multirow}
\usepackage{rotating}
\usepackage[table]{xcolor}

\usepackage[final]{acl}

\usepackage{times}
\usepackage{latexsym}
\usepackage{amsmath} 
\usepackage{tcolorbox}
\usepackage{amsmath, amssymb, amsthm}  
\usepackage{algorithm, algorithmic}    
\usepackage{graphicx}                  
\usepackage{bm}

\usepackage[T1]{fontenc}
\usepackage[utf8]{inputenc}
\usepackage{etoc}

\usepackage{microtype}

\usepackage{inconsolata}

\usepackage{graphicx}

\usepackage{adjustbox}
\usepackage{mdframed}
\usepackage{listings}

\definecolor{myYellow}{RGB}{255, 234, 181} % 淡黄
\definecolor{myGreen}{RGB}{196, 242, 217}  % 淡绿
\definecolor{myBlue}{RGB}{201, 225, 225}   % 淡蓝
\definecolor{Blue2}{RGB}{180, 221, 240}   % 淡蓝
\definecolor{Green2}{RGB}{185, 208, 142}   % 淡蓝
\definecolor{rowgray}{gray}{0.9}

\title{\raisebox{-1.0ex}{\includegraphics[width=0.13\textwidth]{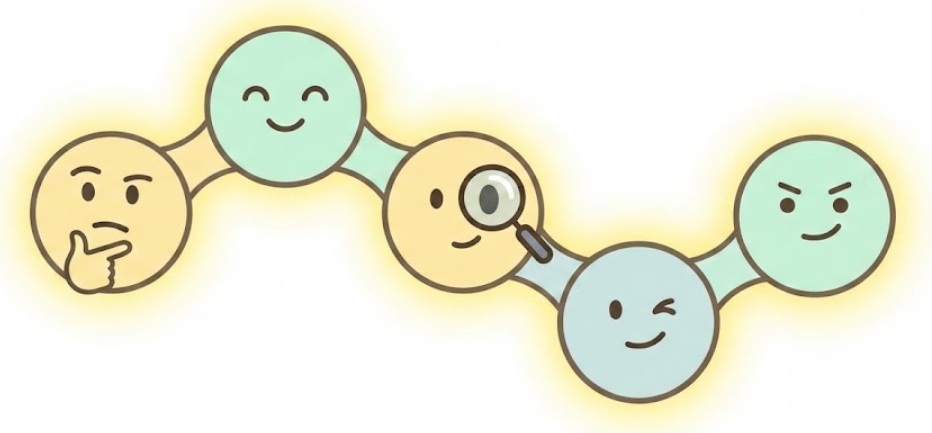}} Neural Chain-of-Thought Search: \\Searching the Optimal Reasoning Path to Enhance Large Language Models}

\author{
 \textbf{Guoming Ling\textsuperscript{1}},
 \textbf{Zhongzhan Huang\textsuperscript{1}},
 \textbf{Yupei Lin\textsuperscript{1}},
 \textbf{Junxin Li\textsuperscript{1}},
 \\
  \textbf{Shanshan Zhong\textsuperscript{1}},
  \textbf{Hefeng Wu\textsuperscript{1}},
   \textbf{Liang Lin\thanks{Corresponding author.}\textsuperscript{1}}
\\
 \textsuperscript{1}Sun Yat-sen University
\\
}

\begin{document}
\etocdepthtag{main}
\maketitle
\begin{abstract}
Chain-of-Thought reasoning has significantly enhanced the problem-solving capabilities of Large Language Models. Unfortunately, current models generate reasoning steps sequentially without foresight, often becoming trapped in suboptimal reasoning paths with redundant steps. In contrast, we introduce Neural Chain-of-Thought Search (NCoTS), a framework that reformulates reasoning as a dynamic search for the optimal thinking strategy. By quantitatively characterizing the solution space, we reveal the existence of sparse superior reasoning paths that are simultaneously more accurate and concise than standard outputs. Our method actively navigates towards these paths by evaluating candidate reasoning operators using a dual-factor heuristic that optimizes for both correctness and computational cost. Consequently, NCoTS achieves a Pareto improvement across diverse reasoning benchmarks, boosting accuracy by over $3.5\%$ while reducing generation length by over $22\%$. 
%We will make our code and data publicly available.
Our code and data are available on \href{https://github.com/MilkThink-Lab/Neural-CoT-Search}{Github}.
\end{abstract}

\section{Introduction}
\label{sec:intro}
% Background & The Problem (Hook -> Gap)
Large Language Models (LLMs) have evolved into specialized Large Reasoning Models (LRMs) \cite{openai2024openaio1card, deepseekai2025deepseekr1incentivizingreasoningcapability, chen2025towards, li2025system, xu2025largereasoningmodelssurvey} that excel at complex tasks through Chain-of-Thought (CoT) reasoning \cite{wei2023chainofthoughtpromptingelicitsreasoning, kojima2022large}. These models achieved state-of-the-art performance on math, logic, and programming benchmarks \cite{zhang2025survey, snell2025scaling}. 
However, recent research indicates that Large Reasoning Models suffer from a strategic bottleneck at reasoning path planning \cite{shojaee2025illusionthinkingunderstandingstrengths, liu2025efficientinferencelargereasoning, sui2025stopoverthinkingsurveyefficient, an2025dontthinklongerthink, jiang2025makes}. They frequently fail to foresee the optimal reasoning direction, causing them to drift into inefficient patterns \cite{kang2025first}. For instance, they may frequently output reflective tokens like "Wait" or "Hmm", triggering unnecessary verification steps or getting stuck in excessive branch exploration\cite{wang2025waitdontneedwait, jiang2025makes, yang2025dynamicearlyexitreasoning}. This behavior suggests a lack of foresight in navigating the reasoning path.

\begin{figure}[t]
  \centering
\includegraphics[width=0.98\linewidth]{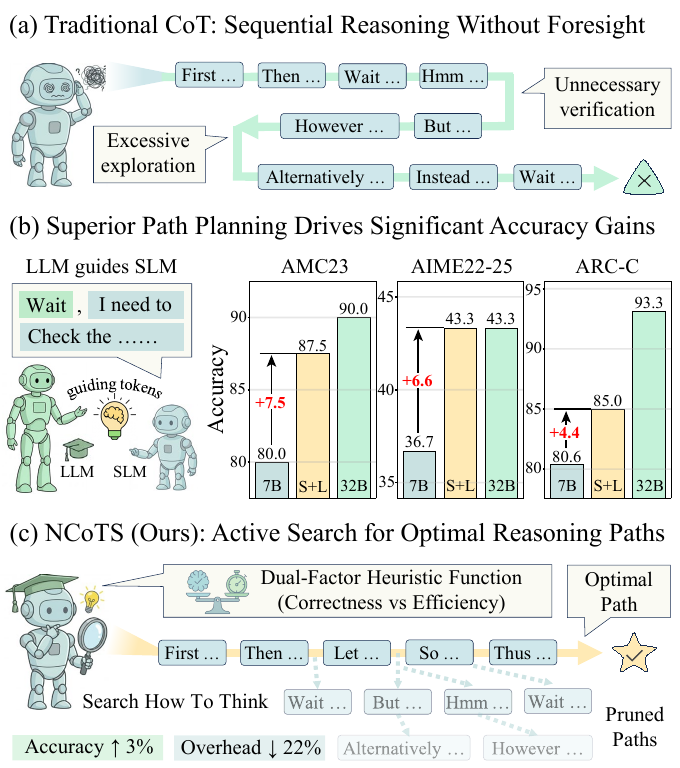}
\vspace{-5pt}
  \caption{Motivation and Overview of our NCoTS.
  (a) Planning Bottleneck in Traditional CoT. 
  (b) Importance of Path Planning. Sparse guiding tokens from a strong teacher significantly boost performance, confirming that path planning is the key bottleneck. 
  (c) The NCoTS framework. Our method reformulates reasoning as a search process, employing a dual-factor heuristic to actively discover paths that are both accurate and concise.}
  \vspace{-15pt}
\label{fig:fig1}
\end{figure}

% Key Insight & Empirical Evidence
We investigate this bottleneck through a hybrid guidance experiment (Detailed in Appendix \ref{appendix:hybird_guidance}). We employed a larger model to generate only the initial token at each reasoning step for a smaller model. These guiding tokens accounted for only $2.9\%$ of the total output but yielded an average accuracy gain of $6.2\%$ across benchmarks (Fig. \ref{fig:fig1}). This result confirms that the core limitation of reasoning models lies in their inability to strategically navigate reasoning paths at critical decision points.

% Solution: Proposed NCoTS to "Search How to Think"
Based on these insights, we propose treating reasoning generation as a dynamic search problem. To validate the potential of this paradigm, we quantitatively characterize the reasoning solution space in Section \ref{sec:solution_space}. This analysis reveals the existence of superior reasoning paths that achieve higher accuracy and lower generation length than standard model outputs. These optimal paths are sparse and difficult to locate via standard sampling, which necessitates a targeted search mechanism to identify them efficiently.
To this end, we introduce Neural Chain-of-Thought Search (NCoTS) in Section \ref{sec:method}. This framework models reasoning as a search for the optimal sequence of reasoning operators. At each decision point, the model evaluates potential directions using a dual-factor heuristic that estimates both correctness and efficiency.
As demonstrated in Section \ref{sec:experiments}, our method actively discovers superior reasoning paths that outperform baselines in both accuracy and efficiency with negligible overhead. We provide a deeper analysis of the proposed framework in Section \ref{sec:futher_discussion}.
We show the related works in Appendix \ref{appendix:related_work}, and summarize the contributions of this paper as follows:

1. We identify the reasoning path planning bottleneck in current reasoning models. Our hybrid guidance experiment reveals that correcting sparse thinking tokens, comprising only $2.9\%$ of the output, yields an average accuracy gain of $6.2\%$.

2. We provide the first quantitative analysis of the reasoning solution space, confirming the existence of superior paths that simultaneously achieve higher accuracy and reduced generation cost.

3. We propose NCoTS, a framework that actively searches how to think to discover superior reasoning paths. NCoTS consistently achieves the highest efficiency metric across all experimental settings, improving average accuracy by over $3.5\%$ while reducing generation length by over $22\%$.

\begin{figure*}[t]
  \centering
  \vspace{-5pt}
\includegraphics[width=0.99\linewidth]{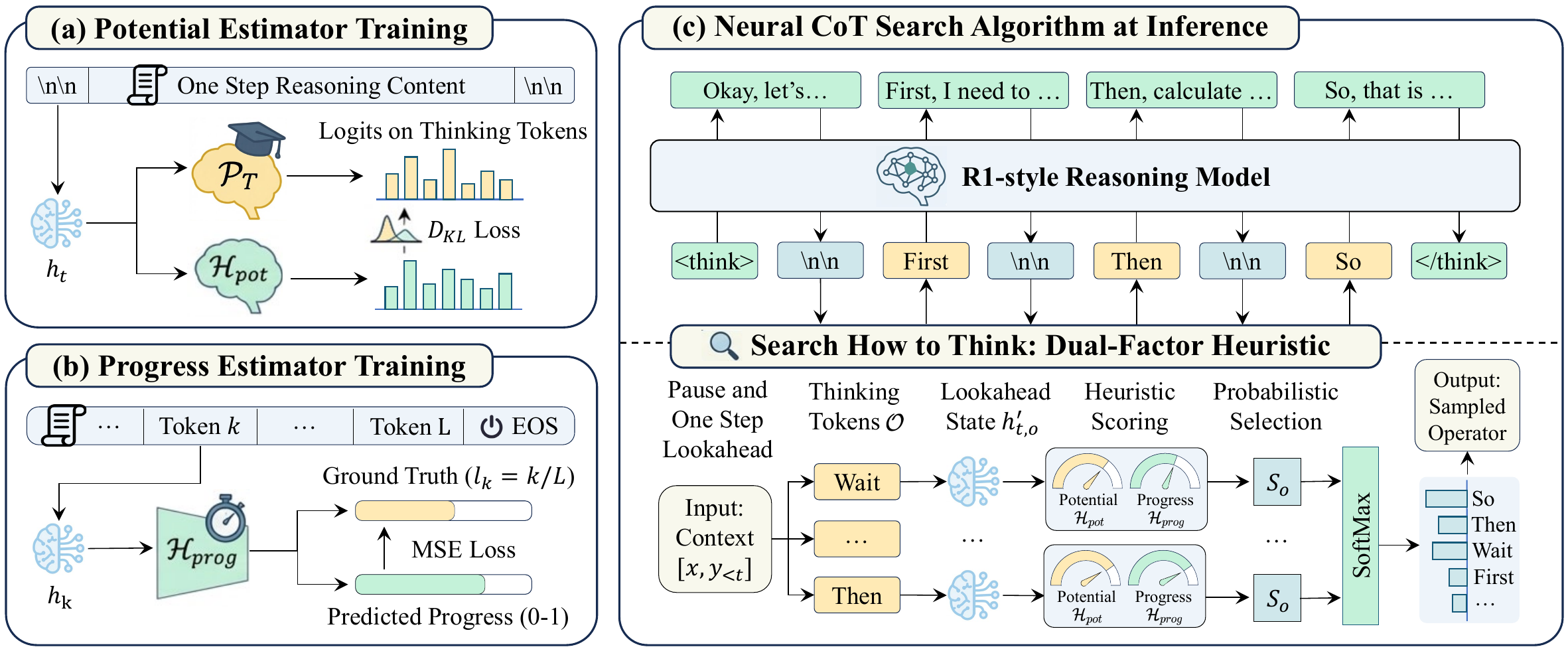}
\vspace{-8pt}
  \caption{Overview of the Neural Chain-of-Thought Search (NCoTS) Framework.
  (a) The Path Potential Estimator employs policy distillation from a teacher model to capture high level planning capabilities.
  (b) The Reasoning Progress Estimator learns to predict the normalized solution progress via token level dense supervision.
  (c) The search algorithm during inference. The model pauses at decision points to \textit{search how to think}. It performs a one step lookahead and evaluates candidate thinking tokens using a dual-factor heuristic function.
  }
  \vspace{-8pt}
\label{fig:framework}
\end{figure*}

\section{Method} 
\label{sec:method} 
We propose Neural Chain-of-Thought Search, a framework that reformulates the generative reasoning process as a dynamic search for the optimal reasoning path. To simultaneously maximize performance and minimize reasoning length, our method explicitly navigates the solution space by evaluating \textit{how to think} at critical decision points.

\subsection{Preliminary}  
\label{sec:preliminary} 
We formalize Chain-of-Thought reasoning as a sequential decision process. Let $x$ denote the input query. The reasoning chain $y$ consists of a sequence of $T$ discrete steps $y = (s_1, s_2, \dots, s_T)$. Each step $s_t$ constitutes a complete semantic unit such as a deduction or calculation. 
Following prior work \cite{yang2025speculativethinkingenhancingsmallmodel}, we mark the completion of a step with a specific delimiter token \verb|"\n\n"|. (See Appendix \ref{appendix:delimiter} for empirical evidence).

We identify the locations of these delimiters as Decision Points. At a given decision point $t$, the model tends to output a thinking token \cite{qian2025demystifyingreasoningdynamicsmutual} to indicate the logical direction of the subsequent step $s_{t+1}$ . For instance, the model might generate "wait" to initiate reflection or "alternatively" to explore other possibilities. We formulate these thinking tokens as Reasoning Operators $o_t$ drawn from a action space $\mathcal{O}$. We define $\mathcal{O}$ as a finite and small set of thinking tokens which allows for efficient enumeration: $\mathcal{O}=\{\text{"Wait"}, \text{"So"}, \text{"Then"}, \text{...}\}$.
The sequence of operators $\alpha = (o_1, o_2, \dots, o_T)$ defines the high-level structure which we term the Reasoning Architecture. Our objective is to find the optimal architecture $\alpha^*$ for a query that maximizes accuracy while minimizing the total sequence length.

\subsection{Overview: Search How to Think} 
\label{sec:overview}  % 写一段 intuition，再写一段方法流程概述
% 完成一步的推理后，不是马上接着进行下一步的推理，而是停下来、仔细去思考下一步该怎么做、该采取什么样的思维模式（think how to think）。采取one step lookahead，想象出这一步的所有可能的思维模式（有限集合且较小）。预测下一步每种方向，导致的正确概率和剩余路径的长度（success rate estimator & cost to end estimator），然后选择综合得分最高的思考方向。通过这样think how to think的方法，减少推理冗余、提升推理性能。
% (1)在关键点停止 (2)想象出所有可能的分支 (3)用双因子启发式函数给每个分支打分 (4)选择分支
\paragraph{Intuition.} 
Existing large reasoning models typically execute reasoning sequentially. Upon completing a step, they immediately generate the subsequent step, often lacking high-level planning. Specifically, the model commits to a specific line of reasoning without evaluating the most effective direction. This lack of foresight may trap models in suboptimal paths, leading to redundant verification loops or verbose derivations.

\paragraph{Proposed Mechanism.} To address this, we introduce a mechanism to \textit{search how to think}. 
Fig. \ref{fig:framework} illustrates the overview of our framework, which comprises four phases.
\underline{(1) Pause Generation}: The standard generation halts immediately upon detecting a step delimiter. 
\underline{(2) Lookahead Simulation}: The model simulates potential reasoning directions by projecting all candidate operators from the set $\mathcal{O}$ into the future context. 
\underline{(3) Heuristic Evaluation}: A dual-factor heuristic function assesses each direction by estimating its success probability and computational cost. 
\underline{(4) Strategic Selection}: The model samples the optimal operator based on these estimates and resumes generation. This active decision process prunes inefficient branches before they consume computational resources.

\subsection{Dual-Factor Heuristic Function}  
\label{sec:heuristics}

We employ a composite heuristic function $\mathcal{H}(h_t, o)$ to evaluate the efficacy of applying operator $o$ at the current hidden state $h_t$. This function comprises two specialized estimators designed to quantify the quality and efficiency of the reasoning path.

\paragraph{Path Potential Estimator.}
The first component is the Path Potential Estimator $\mathcal{H}_{\text{pot}}$. It predicts the probability that a specific reasoning direction will lead to a correct solution. We implement this estimator as a linear projection layer taking the final hidden state as input to output logits over the operator set $\mathcal{O}$. As demonstrated in Section \ref{sec:intro}, Larger Models possess stronger capabilities in high-level planning. Therefore, we train this estimator via policy distillation from a fixed Teacher LRM. We treat the teacher's probability distribution over $\mathcal{O}$ as the expert policy $P_T$. The estimator is optimized by minimizing the Kullback-Leibler divergence:
\begin{equation}
\mathcal{L}_{\text{pot}} = \mathbb{E}_{h_t \sim \mathcal{D}} \left[ D_{\text{KL}}\Big( P_T(h_t) \;\big\|\; \mathcal{H}_{\text{pot}}(h_t) \Big) \right].
\end{equation}
This estimator effectively transfers the strategic planning capabilities of the teacher into the search process, serving as the compass for correctness. 

\paragraph{Reasoning Progress Estimator.}
The second component is the Reasoning Progress Estimator $\mathcal{H}_{\text{prog}}$. It estimates the efficiency of a reasoning path. We implement this estimator as a linear regression head that maps the hidden state to a scalar value representing normalized progress. Similar to recent works on reasoning monitoring \cite{eisenstadt2025overclockingllmreasoningmonitoring}, this estimator predicts the completion ratio of the solution given the current state. We train this estimator on a token-level dense supervision task. For each training query, we collect multiple complete reasoning paths. Specifically, for every token at index $k$ within a completed path of total length $L$, we construct a training pair $(h_k, l_k)$. Here, $h_k$ denotes the hidden state and $l_k=k/L$ represents the ground truth normalized progress, indicating the portion of the solution completed. The estimator $\mathcal{H}_{\text{prog}}$ projects $h_k$ to a scalar value, trained by minimizing the Mean Squared Error:
\begin{equation}
\mathcal{L}_{\text{prog}} = \mathbb{E}_{(h_k, l_k) \sim \mathcal{D}} \left[ \left\| \mathcal{H}_{\text{prog}}(h_k) - l_k \right\|^2 \right].
\end{equation}
By maximizing this estimated progress, the search algorithm favors operators that significantly advance the reasoning state toward the solution, effectively penalizing verbose or circular steps.

\subsection{Search Algorithm} 
\label{sec:search_algorithm}
We optimize the reasoning path by actively \textit{searching how to think} during inference. This strategy evaluates potential reasoning directions at decision points to identify the optimal path.

\paragraph{One Step Lookahead.} 
At decision point $t$, marked by the delimiter \verb|"\n\n"|, we proactively explore the potential future space. Let $y_{<t}$ denote the current reasoning path. For each candidate operator $o \in \mathcal{O}$, we simulate the next step by appending $o$ to the KV cache of model $\mathcal{M}$:
\begin{equation}
\mathbf{h}'_{t,o} = \mathcal{M}\big( [x, y_{<t}, o] \big), \quad \forall o \in \mathcal{O}.
\end{equation}

This yields the lookahead hidden state $\mathbf{h}'_{t,o}$. Given that the thinking token governs the thinking mode as detailed in Section \ref{sec:futher_discussion}, this lightweight lookahead captures the semantic trajectory of the branch without the overhead of full step generation.

\paragraph{Heuristic Scoring.} 
Once the lookahead states are generated, we assign a composite score $S(o)$ to each branch by aggregating the outputs of the dual-factor heuristics. The score integrates both the potential for accuracy and the efficiency of progress:
\begin{equation}
S(o) = \underbrace{\mathcal{H}_{\text{potential}}(h_t, o)}_{\text{Success Potential}} + \lambda \cdot \underbrace{\mathcal{H}_{\text{progress}}(h'_{t,o})}_{\text{Efficiency Progress}}.
\label{eq:scoring}
\end{equation}
Here, $\lambda$ is a hyperparameter that governs the emphasis on conciseness. A higher $\lambda$ encourages the model to select more concise reasoning paths.

\paragraph{Probabilistic Selection.} 
To ensure diversity and avoid local optima, we convert these scores into a probabilistic search policy $P_{\text{search}}$ using Softmax function with a temperature parameter $\tau$:
\begin{equation}
P_{\text{search}}(o|h_t) = \frac{\exp\left( S(o) / \tau \right)}{\sum_{o' \in \mathcal{O}} \exp\left( S(o') / \tau \right)}.
\end{equation}
The final operator is selected by sampling $o^* \sim P_{\text{search}}$. This procedure ensures that the selected reasoning direction is both strategically sound and computationally efficient.

\begin{figure*}[t]
  \centering
  \vspace{-5pt}
\includegraphics[width=0.99\linewidth]{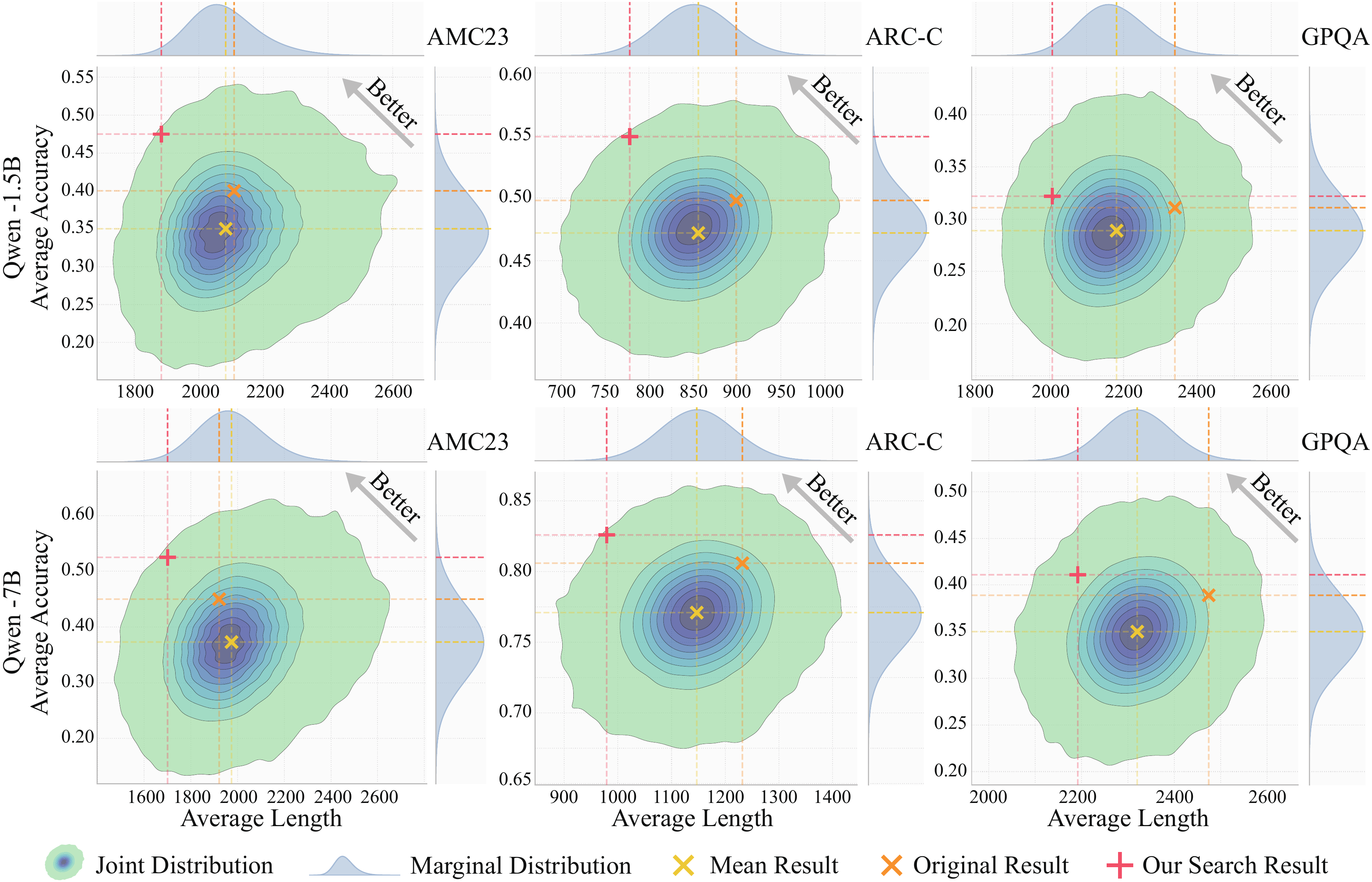}
\vspace{-5pt}
  \caption{Visualization of the reasoning solution space. The region to the upper-left of the Original result indicates the existence of superior solutions. This confirms that paths with higher accuracy and lower length are attainable, validating the feasibility of our search framework. The red cross mark represents our method, demonstrating that our strategy successfully discovers these superior paths that optimize both accuracy and conciseness.}
  \vspace{-5pt}
\label{fig:distribution}
\end{figure*}

\section{Experiments}
\label{sec:experiments}
In this section, we empirically validate the proposed framework. We first characterize the reasoning solution space, confirming the existence of superior paths that achieve higher accuracy and lower length than standard generation. We then demonstrate that our method actively locates these paths, consistently achieving the highest efficiency metrics ($\eta$) across all experimental settings.

\subsection{Experimental Setup}
\label{experiment_setup}
\paragraph{Datasets.} We evaluate the performance of our method across four diverse benchmarks. The selected benchmarks include AMC23, ARC-C \cite{clark2018thinksolvedquestionanswering}, GPQA \cite{rein2023gpqagraduatelevelgoogleproofqa}, and GSM8K \cite{cobbe2021trainingverifierssolvemath} which collectively cover symbolic deductive reasoning, commonsense reasoning, expert knowledge reasoning and multi-step arithmetic reasoning. We provide details of these benchmarks in Appendix \ref{appendix:benchmarks}.

\paragraph{Models.} To broadly explore the characteristics of the solution space, our analysis in Section \ref{sec:solution_space} employs multiple models of varying sizes and architectures: DeepSeek-R1-Distill-Qwen-\{1.5B, 7B, 14B, 32B\}, and DeepSeek-R1-Distill-Llama-8B \cite{deepseekai2025deepseekr1incentivizingreasoningcapability, qwen2025qwen25technicalreport, dubey2024llama}. For the evaluation of our search method (Section \ref{sec:result_analysis}), we employ two configurations: a small pair, which uses \{SLM=Qwen-1.5B, LLM=Qwen-7B\}, and a large pair, which uses \{SLM=Qwen-7B, LLM=Qwen-32B\}.

\paragraph{Baselines.} We compare the proposed framework against six baselines. We use Mean and Original to represent the performance of standard sampling. The evaluation also includes recent strategies for optimizing reasoning efficiency such as NoWait \cite{wang2025waitdontneedwait}, AdaptThink \cite{zhang2025adaptthinkreasoningmodelslearn}, ThinkPrune \cite{hou2025thinkprunepruninglongchainofthought} and Laser \cite{liu2025learnreasonefficientlyadaptive}. See Appendix \ref{appendix:baselines} for details.

\paragraph{Metrics.} We report task-specific Accuracy ($A$) and the average token count ($L$). To quantify the trade-off between performance gains and computational cost, we adopt a composite Efficiency Metric ($\eta$), inspired by previous works on efficient reasoning \cite{an2025dontthinklongerthink, qu2025surveyefficientreasoninglarge}. This metric places a quadratic emphasis on accuracy, as computational savings are secondary to correctness:
\begin{equation}\eta = \underbrace{\left( \frac{\mathbb{E}_{y \sim \pi^*} [A(y)]}{\mathbb{E}_{y_0 \sim \pi} [A(y_0)]} \right)^2}_{\text{Performance Gain}} \cdot \underbrace{\frac{\mathbb{E}_{y_0 \sim \pi} [L(y_0)]}{\mathbb{E}_{y \sim \pi^*} [L(y)]}}_{\text{Computational Savings}}.\label{eq:efficiency_metric}\end{equation}
Here, $\pi^*$ denotes our search-augmented policy and $\pi$ represents the original model. $A(\cdot)$ measures solution correctness and $L(\cdot)$ denotes sequence length. A value of $\eta > 1$ indicates that the method improves the reasoning density and provides more correct reasoning per unit of computation.

\subsection{Implementation Details}
\label{sec:imple_details}
\paragraph{Characterization of Solution Space.} 

We conduct a randomized search experiment to characterize the architectural search space $\mathcal{A}$ and empirically map the performance boundaries of the model. For each query, we generate multiple independent reasoning paths by intervening at every step delimiter to sample the reasoning operator $o_t$ from the uniform distribution over $\mathcal{O}$.
We then aggregate these paths to construct density heatmaps on the Accuracy versus Length plane. This visualization reveals the distribution of potential search strategies and the theoretical limits of the model. Appendix \ref{appendix:random_search} provides more details regarding the experimental setup and searching mechanism. 
% Fig. \ref{fig:distribution} shows the experimental results and Section \ref{sec:solution_space} provides detailed analysis.

\paragraph{Training.} We initialize the potential estimator using weights from the pre-trained language head of the student model. Specifically, we extract the rows of its embedding matrix corresponding to the thinking tokens in $\mathcal{O}$ to preserve the model's initial semantic priors. The progress estimator is initialized randomly. Training is performed on a composite dataset comprising LogicQA \cite{liu2020logiqachallengedatasetmachine}, Math500 \cite{hendrycks2021measuringmathematicalproblemsolving}, AIME22-25 \cite{balunovic_srimatharena_2025}, and HumanEval \cite{chen2021evaluatinglargelanguagemodels}. For the potential estimator, we employ a distillation objective. Given a query, the model generates steps until a decision point is reached. We then compute logits for the thinking tokens using the fixed Teacher LLM and minimize the KL divergence between the Teacher's distribution and the estimator's output. The progress estimator is trained via Mean Squared Error to predict the complete ratio of solution, as detailed in Section \ref{sec:heuristics}.

\paragraph{Testing.} For all methods, we set the temperature to $0.6$, top-p to $0.95$, and the global maximum token limit to $4096$. For our search method, we impose a maximum limit of $50$ reasoning steps. During inference, we set the balancing hyperparameter $\lambda=1$ and sample the reasoning operators based on the composite score $S(o)$, following the search policy detailed in Section \ref{sec:search_algorithm}. The prompt used is: 
\verb|Please reason step by step,|
\verb|and put your final answer within \boxed{}|
, following previous works \cite{chen2025verithinkerlearningverifymakes, yang2025speculativethinkingenhancingsmallmodel, cheng2025revisitingchainofthoughtpromptingzeroshot}.

\begin{table*}[!t]
\centering
\resizebox{\linewidth}{!}{
    \setlength{\tabcolsep}{4pt} 
    \begin{tabular}{l|ccc|ccc|ccc|ccc|ccc}
    \toprule
        & \multicolumn{3}{c|}{AMC23} & \multicolumn{3}{c|}{ARC-C} & \multicolumn{3}{c|}{GPQA} & \multicolumn{3}{c|}{GSM8K} & \multicolumn{3}{c}{Average} \\
    \multirow{-2}{*}{Method} & Acc$\uparrow$ & Length$\downarrow$ & $\bm{\eta}$$\uparrow$ & Acc$\uparrow$ & Length$\downarrow$ & $\bm{\eta}$$\uparrow$ & Acc$\uparrow$ & Length$\downarrow$ & $\bm{\eta}$$\uparrow$ & Acc$\uparrow$ & Length$\downarrow$ & $\bm{\eta}$$\uparrow$ & $\Delta$Acc$\uparrow$ & $\Delta$Length$\downarrow$ & $\bm{\eta}$$\uparrow$ \\ \midrule
    
    % --- 1.5B Section ---
    \multicolumn{16}{l}{\textit{DeepSeek-R1-Distill-Qwen-1.5B}} \\ \midrule
    
    {\color[HTML]{7F7F7F} Mean} 
    & {\color[HTML]{7F7F7F} 35.0} & {\color[HTML]{7F7F7F} 2083} & {\color[HTML]{7F7F7F} 0.775} 
    & {\color[HTML]{7F7F7F} 47.2} & {\color[HTML]{7F7F7F} 856} & {\color[HTML]{7F7F7F} 0.943} 
    & {\color[HTML]{7F7F7F} 28.9} & {\color[HTML]{7F7F7F} 2181} & {\color[HTML]{7F7F7F} 0.926} 
    & {\color[HTML]{7F7F7F} 80.9} & {\color[HTML]{7F7F7F} 1846} & {\color[HTML]{7F7F7F} 0.997} 
    & {\color[HTML]{7F7F7F} -3.0} & {\color[HTML]{7F7F7F} -4.4\%} & {\color[HTML]{7F7F7F} 0.910} \\
    
    {\color[HTML]{7F7F7F} Original} 
    & {\color[HTML]{7F7F7F} 40.0} & {\color[HTML]{7F7F7F} 2109} & {\color[HTML]{7F7F7F} 1.000} 
    & {\color[HTML]{7F7F7F} 49.8} & {\color[HTML]{7F7F7F} 899}  & {\color[HTML]{7F7F7F} 1.000} 
    & {\color[HTML]{7F7F7F} 31.1} & {\color[HTML]{7F7F7F} 2339}  & {\color[HTML]{7F7F7F} 1.000} 
    & {\color[HTML]{7F7F7F} 83.0} & {\color[HTML]{7F7F7F} 1938}  & {\color[HTML]{7F7F7F} 1.000}
    & {\color[HTML]{7F7F7F} +0.0} & {\color[HTML]{7F7F7F} +0.0\%} & {\color[HTML]{7F7F7F} 1.000} \\
    
    NoWait 
    & 40.0 & 1967 & 1.072 & 50.8 & 812 & 1.152 & 28.9 & 1992 & 1.014 & 84.1 & 1211 & 1.641 & +0.0 & -17.2\% & 1.220 \\
    
    AdaptThink 
    & 42.5 & 1926 & 1.236 & 50.5 & 897 & 1.031 & 32.2 & 2429 & 1.032 & 85.4 & 1109 & 1.848 & +1.7 & -12.0\% & 1.287 \\
    
    ThinkPrune 
    & 45.0 & 1803 & 1.480 & 51.5 & 768 & 1.252 & 31.1 & 1900 & 1.231 & 84.9 & 1191 & 1.702 & +2.1 & -21.6\% & 1.416 \\
    
    Laser 
    & 40.0 & 1902 & 1.109 & 50.8 & 870 & 1.075 & 27.8 & 2121 & 0.881 & 82.3 & 1064 & 1.792 & -0.7 & -16.9\% & 1.214 \\
    
    Ours 
    & 47.5 & 1884 & \textbf{1.578} & 54.9 & 778 & \textbf{1.405} & 32.2 & 2007 & \textbf{1.249} & 85.4 & 954 & \textbf{2.148} & \textbf{+4.0} & \textbf{-22.3\%} & \textbf{1.595} \\ \midrule
    
    % --- 7B Section ---
    \multicolumn{16}{l}{\textit{DeepSeek-R1-Distill-Qwen-7B}} \\ \midrule
    
    {\color[HTML]{7F7F7F} Mean} 
    & {\color[HTML]{7F7F7F} 37.3} & {\color[HTML]{7F7F7F} 1974} & {\color[HTML]{7F7F7F} 0.669} 
    & {\color[HTML]{7F7F7F} 77.1} & {\color[HTML]{7F7F7F} 1147} & {\color[HTML]{7F7F7F} 0.983} 
    & {\color[HTML]{7F7F7F} 35.0} & {\color[HTML]{7F7F7F} 2320} & {\color[HTML]{7F7F7F} 0.864} 
    & {\color[HTML]{7F7F7F} 88.1} & {\color[HTML]{7F7F7F} 1649} & {\color[HTML]{7F7F7F} 0.976} 
    & {\color[HTML]{7F7F7F} -4.3} & {\color[HTML]{7F7F7F} -3.2\%} & {\color[HTML]{7F7F7F} 0.873} \\
    
    {\color[HTML]{7F7F7F} Original} 
    & {\color[HTML]{7F7F7F} 45.0} & {\color[HTML]{7F7F7F} 1921} & {\color[HTML]{7F7F7F} 1.000} 
    & {\color[HTML]{7F7F7F} 80.6} & {\color[HTML]{7F7F7F} 1232} & {\color[HTML]{7F7F7F} 1.000} 
    & {\color[HTML]{7F7F7F} 38.9} & {\color[HTML]{7F7F7F} 2474} & {\color[HTML]{7F7F7F} 1.000} 
    & {\color[HTML]{7F7F7F} 90.3} & {\color[HTML]{7F7F7F} 1690}  & {\color[HTML]{7F7F7F} 1.000} 
    & {\color[HTML]{7F7F7F} +0.0} & {\color[HTML]{7F7F7F} +0.0\%} & {\color[HTML]{7F7F7F} 1.000} \\
    
    NoWait 
    & 50.0 & 1894 & 1.252 & 80.3 & 1082 & 1.130 & 38.9 & 2248 & 1.101 & 91.3 & 1147 & 1.506 & +1.4 & -13.7\% & 1.247 \\
    
    AdaptThink 
    & 47.5 & 1910 & 1.121 & 82.9 & 1088 & 1.198 & 42.2 & 2393 & 1.218 & 92.6 & 1086 & 1.635 & +2.6 & -12.8\% & 1.293 \\
    
    Laser 
    & 50.0 & 1650 & 1.437 & 79.9 & 944 & 1.283 & 38.9 & 2279 & 1.086 & 93.0 & 968 & 1.852 & +1.8 & -22.0\% & 1.414 \\
    
    Ours 
    & 52.5 & 1700 & \textbf{1.538} & 82.6 & 979 & \textbf{1.322} & 41.1 & 2192 & \textbf{1.261} & 92.6 & 899 & \textbf{1.976} & \textbf{+3.5} & \textbf{-22.6\%} & \textbf{1.524} \\ \bottomrule
    \end{tabular} 
}
\caption{Main results comparing the proposed Neural CoT Search against baselines on AMC23, ARC-C, GPQA, and GSM8K benchmarks. The table reports task-specific Accuracy (Acc), Average Generation Length (Length), and the Efficiency Metric ($\bm{\eta}$). Our method consistently achieves the highest $\bm{\eta}$ across all settings, demonstrating simultaneous improvements in accuracy and efficiency. Best results are highlighted in bold.}
\label{tab:main_results}
\end{table*}

\subsection{The Reasoning Solution Space}
\label{sec:solution_space}
Fig. \ref{fig:distribution} presents the density heatmaps of Average Length versus Average Accuracy derived from our random search characterization (See Appendix \ref{appendix:more_result} for more results). This visualization reveals four insights into the nature of CoT reasoning: 

\underline{(1) Operator Choice Drives High Variance.} The reasoning path is highly sensitive to the choice of reasoning operators. Selecting different operators leads to vastly different outcomes in both accuracy and length. This structural divergence confirms that the high-level planning of the reasoning path is a critical determinant of the final solution quality.
\underline{(2) Suboptimality of Standard Decoding.} The Original baseline consistently outperforms the random Mean baseline but remains far from the theoretical performance boundary. This gap suggests that the model's standard generation strategy fails to exploit the full intrinsic potential of the model.
\underline{(3) Existence of Superior Paths.} The heatmaps reveal a region in the upper-left quadrant with higher accuracy and lower length than the original baseline. These Pareto-superior solutions are empirical proof that it is feasible to simultaneously optimize correctness and cost.
\underline{(4) Sparsity of Superior Solutions.} The region containing these superior paths is extremely sparse compared to the dense clusters of suboptimal paths. This sparsity explains why standard sampling fails to yield consistent improvements. The probability of randomly encountering a superior path is negligible, necessitating a targeted search approach.

\subsection{Efficacy of the Proposed Search Strategy}
\label{sec:result_analysis}

Table \ref{tab:main_results} compares our method against baselines across DeepSeek-R1-Distill-Qwen-1.5B and DeepSeek-R1-Distill-Qwen-7B. 
While many existing baselines struggle to balance the trade-off between performance and cost, our method simultaneously enhances accuracy and reduces computational cost. 
On the 1.5B model, we achieve a $4.0\%$ accuracy gain and a $22.3\%$ reduction in token usage. 
On the 7B model, NCoTS improves average accuracy by $3.5\%$ and decreases generation length by $22.6\%$. 
Notably, on GSM8K with the 1.5B model, our approach reduces the generation length by over $50\%$ while achieving an accuracy gain of $2.4\%$.
Moreover, on GSM8K with the 7B model, accuracy improves by $2.3\%$ with a length reduction of $47\%$, and on AMC23 accuracy improves substantially by $7.5\%$ with length reduced by $12\%$.
Our method consistently achieves the highest efficiency metric $\bm{\eta}$ across all settings, yielding an average $\bm{\eta}$ of $1.595$ for the 1.5B model and $1.524$ for the 7B model. This confirms that our search strategy maximizes reasoning density and effectively prunes redundant steps to deliver more correct reasoning per unit of computation.

Furthermore, we observe a distinct correlation between the nature of the task and the magnitude of efficiency gains. The method excels in reasoning-intensive tasks. On GSM8K and AMC23, it achieves the highest efficiency scores between $1.5$ and $2.1$, as the search mechanism effectively navigates complex reasoning branches. In hybrid tasks like ARC-C, which require a blend of common sense and reasoning, gains remain substantial with $\bm{\eta}$ ranging from $1.3$ to $1.4$. On knowledge-intensive tasks such as GPQA, efficiency gains are the lowest at approximately $1.2$. This is expected behavior, as performance in these domains relies more on factual retrieval than strategic planning, yet the consistent improvement across all benchmarks validates the generalizability of our framework.

\section{Further Discussion}
\label{sec:futher_discussion}
In this Section, We conduct a more comprehensive analysis of the proposed search framework. For more analysis, please refer to Appendix \ref{appendix:thinking_tokens} and \ref{appendix:design_choice}.

\begin{figure}[t]
  \centering
\includegraphics[width=0.98\linewidth]{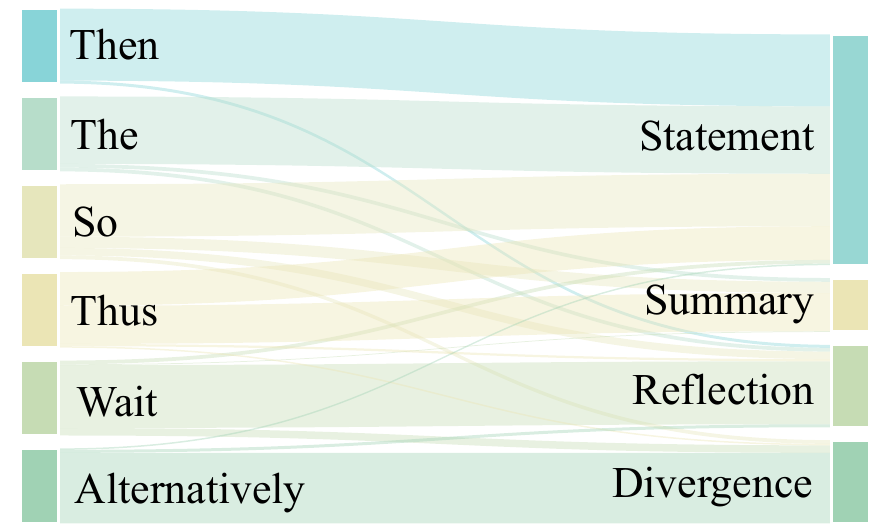}
\vspace{-5pt}
  \caption{Correlation between thinking tokens and thinking modes. This Sankey diagram illustrates the strong influence of the chosen operator (thinking token) on the functional purpose of the subsequent reasoning step.}
  \vspace{-15pt}
\label{fig:sankey}
\end{figure}

\begin{mdframed}[backgroundcolor=gray!8]
\begin{minipage}{\linewidth}
(1) How does the thinking token affect the corresponding reasoning step?
\end{minipage}
\end{mdframed}
In section \ref{sec:method}, we use thinking tokens from the operator set $\mathcal{O}$ to steer the reasoning direction at each decision point. To illustrate the influence of these tokens, we analyzed a large corpus of reasoning paths generated by DeepSeek-R1-Distill-Qwen-1.5B on the AMC23 benchmark. We extracted each (operator, step) pair and employed DeepSeek-V3 to classify the functional purpose of the step $s_i$ into one of four modes: Statement, Summary, Reflection, or Divergence (see Appendix \ref{appendix:thinking_tokens} for prompt, methodology and more results).
As shown in Fig. \ref{fig:sankey}, our analysis reveals a strong correspondence between the chosen operator and the resulting thinking mode. For instance, the "Wait" operator consistently precedes Reflection steps, whereas "Then" strongly correlates with Statement steps.

Psychological studies suggest that human System 2 reasoning involves multiple distinct modes of thinking, such as stating, summarizing, reflecting, and exploring \cite{evans2008dual, moshman2014epistemic}. People dynamically switch between them during complex reasoning. We argue that for LLMs to solve complex problems, they also require this ability to dynamically shift their thinking mode. 
A key insight of our work is that thinking tokens are not just superficial prefixes, they function as a control mechanism to select the thinking mode for next step. 
Leveraging this insight, our method dynamically guides the model's thinking modes, thereby steering the reasoning path toward a better solution.

\begin{figure}[t]
  \centering
\includegraphics[width=0.98\linewidth]{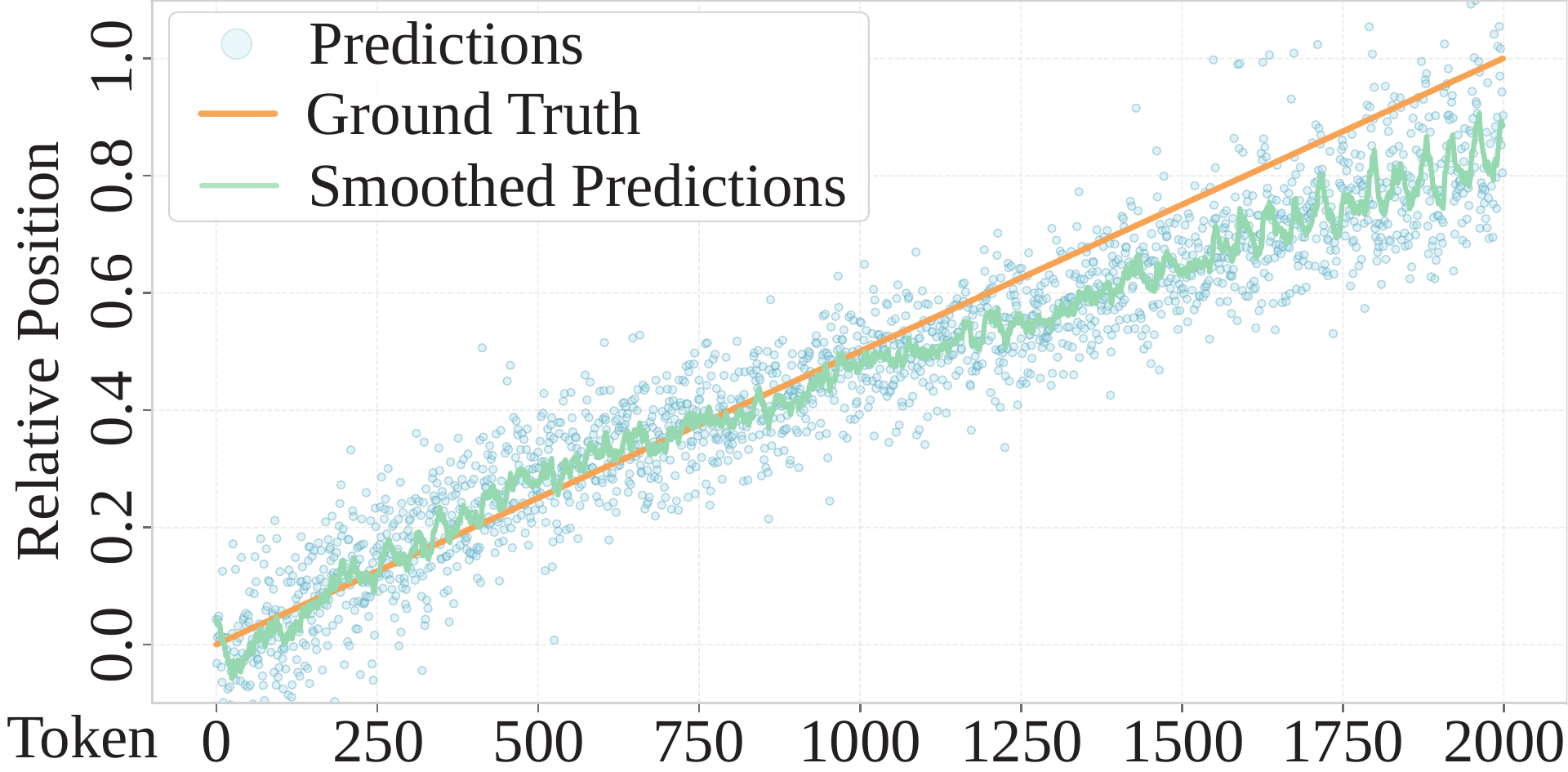}
\vspace{-5pt}
  \caption{
  A comparison of the estimated progress against the ground truth progress. The exponentially smoothed estimator output closely aligns with the ground truth progress $y=x/L$.
  }
  \vspace{-12pt}
\label{fig:len_pred}
\end{figure}

\begin{mdframed}[backgroundcolor=gray!8]
\begin{minipage}{\linewidth}
(2) Does the reasoning progress estimator predict the progress accurately?
\end{minipage}
\end{mdframed}
We introduce the reasoning progress estimator $\mathcal{H}_{\text{prog}}$ in Section \ref{sec:heuristics}, grounded in recent evidence that the hidden states of reasoning models implicitly encode the progress of the solution \cite{eisenstadt2025overclockingllmreasoningmonitoring}.
Figure \ref{fig:len_pred} plots the estimator's predictions against ground-truth normalized positions.
The exponentially smoothed prediction trajectory aligns well with the true progress, demonstrating that a lightweight regression estimator effectively extracts this signal and estimate the remaining computational cost. 
The visible variance in the scatter plot reflects semantic sensitivity rather than stochastic noise. 
As noted in \cite{eisenstadt2025overclockingllmreasoningmonitoring}, reflective tokens (e.g., ``Wait'', ``Hmm'') induce drops in predicted progress, correctly signaling reasoning expansion, while decisive operators (e.g., ``Therefore'') indicate proximity to the solution.
In our search framework, we prioritize the capability to distinguish efficiency over exact progress prediction. The reasoning progress estimator need only preserve the correct preference ordering by assigning higher values to efficient operators (e.g., $v_{\text{``Then''}} > v_{\text{``Wait''}}$). This ensures that the search algorithm correctly prioritizes more efficient branches without necessitating precise estimation of the absolute length.

\begin{mdframed}[backgroundcolor=gray!8]
\begin{minipage}{\linewidth}
(3) Is the path potential estimator or the reasoning progress estimator necessary?
\end{minipage}
\end{mdframed}
% 消融
To validate our dual-factor heuristic design, we conducted an ablation study by removing the potential and progress estimators respectively. Table \ref{tab:ablation} reports the results on the English Math Competition subset of OlympiadBench \cite{he2024olympiadbench}. The results demonstrate that each component contributes to the search process in a unique and indispensable way. 
The configuration without the progress estimator achieves high accuracy but fails to maximize efficiency, as the potential estimator prioritizes correctness without incentives to prune valid but redundant steps. 
Conversely, removing the potential estimator leads to a collapse in performance. It is worth noting that while this setting reduces length compared to the original baseline, it is less efficient than our full method. This occurs because the progress estimator, lacking semantic guidance, tends to select operators that disrupt the logical flow, causing the model to generate incoherent, compensatory text in an attempt to recover.
Therefore, our framework relies on the synergy between the two estimators. The potential estimator leverages distilled strategic priors to identify paths with high success probability. The progress estimator proactively steers the search toward the most compact reasoning paths. This combination steers the search toward reasoning paths that are simultaneously correct and concise.

\begin{table}[t]
  \centering
    \begin{tabular}{lrrr}
    \toprule
    \multicolumn{4}{c}{OlympiadBench} \\
    \midrule
    Method & \multicolumn{1}{c}{Acc $\uparrow$} & \multicolumn{1}{c}{Len $\downarrow$} & \multicolumn{1}{c}{$\eta$ $\uparrow$} \\
    \midrule
    Original & 51.3 & 8765   & 1.000  \\
    Ours w/o $\mathcal{H}_{potential}$ & 48.7 & 6334  & 1.247  \\
    Ours w/o $\mathcal{H}_{progress}$  & 53.0 & 5562   & 1.682  \\
    Ours  & 52.2 & 4187   & 2.167  \\
    \bottomrule
    \end{tabular}
  \caption{Ablation study of the dual-factor heuristic. The results verify that both potential and progress estimators are essential for balancing correctness and conciseness.}
  \label{tab:ablation}
  \vspace{-10pt}
\end{table}%

\begin{mdframed}[backgroundcolor=gray!8]
\begin{minipage}{\linewidth}
(4) Is the search paradigm we proposed compatible with other methods?
\end{minipage}
\end{mdframed}
Our proposed search paradigm is compatible with existing methods. Since our approach operates at the decoding stage by intervening in the selection of reasoning operators, it functions as a plug-and-play module that is orthogonal to model architecture modifications or sample-level routing strategies.
To demonstrate this compatibility, we analyze the integration of our method with AdaptThink \cite{zhang2025adaptthinkreasoningmodelslearn}. AdaptThink represents a class of long/short thinking strategies that dynamically determine the inference budget based on the difficulty of the input query. The synergy is clear: AdaptThink optimizes the macro-level resource allocation (deciding when to reason), while our method optimizes the micro-level reasoning path (steering how to reason). As shown in Table \ref{tab:compatible}, the composite method achieves additive efficiency gains, proving that our search effectively complements budget-adaptive baselines.

\begin{mdframed}[backgroundcolor=gray!8]
\begin{minipage}{\linewidth}
(5) How is the cost and latency of the dual-factor heuristic function?
\end{minipage}
\end{mdframed}
The overhead introduced by our dual-factor heuristic function is negligible in terms of both memory and latency. 
Regarding parameter efficiency, for the 1.5B model (hidden dimension $d=1536$), the potential estimator ($d \to \left| \mathcal{O} \right|$) and progress estimator ($d \to 1$) collectively introduce approximately $2.6 \times 10^4$ parameters. This represents a mere $0.0017\%$ increase, incurring negligible memory overhead. 
Inference latency is mitigated by sparse activation and parallel lookahead. The search mechanism activates strictly at critical decision points, which comprise only $3\%$ of total tokens, allowing the model to execute standard decoding for the remaining $97\%$. When activated, candidate branches share an identical prefix, enabling us to compute lookahead steps in a single parallel batch via KV caching.
Ultimately, this minor cost is surpassed by efficiency gains; our method reduces average generation length by over $22\%$ across all benchmarks. This substantial decrease in generation length yields a net reduction in aggregate computational operations.

\begin{table}[t]
  \centering
    \begin{tabular}{lrrr}
    \toprule
    \multicolumn{4}{c}{OlympiadBench} \\
    \midrule
    Method & \multicolumn{1}{c}{Acc $\uparrow$} & \multicolumn{1}{c}{Len $\downarrow$} & \multicolumn{1}{c}{$\eta$ $\uparrow$} \\
    \midrule
    Original & 51.3 & 8765   & 1.000  \\
    AdaptThink & 52.2 & 5267   & 1.723  \\
    Ours  & 52.2 & 4187   & 2.167  \\
    Ours + AdaptThink & 54.8 & 3691   & 2.708  \\
    \bottomrule
    \end{tabular}
  \caption{Compatibility analysis with AdaptThink. The results demonstrate that our method complements existing strategies to achieve additive efficiency gains.}
  \label{tab:compatible}
  \vspace{-8pt}
\end{table}%

\section{Conclusion}
In this paper, we introduce NCoTS, a framework that searches for optimal reasoning paths by dynamically steering the thinking modes at decision points. By explicitly optimizing for correctness and conciseness with a dual-factor heuristic, NCoTS achieves a Pareto improvement, boosting accuracy by over $3.5\%$ while reducing generation length by $22\%$. Our findings demonstrate that the bottleneck of efficient reasoning lies in the myopia of next-token prediction; resolving this requires equipping models with the foresight to plan \textit{how to think}.

\section*{Limitations}
We propose a search mechanism guided by a defined operator set. 
However, our current set is primarily optimized for English STEM reasoning and does not account for other languages or creative tasks. 
Fortunately, the framework allows for straightforward extension to multilingual or creative domains by recalibrating these thinking tokens.
Additionally, while our potential estimator relies on teacher supervision which theoretically bounds the planning capability, future works could employ reinforcement learning to enable self-improved exploration beyond the teacher's distribution. 
Furthermore, our reliance on static newline delimiters effectively captures major pauses but may be too rigid for non-standard formats, suggesting a need for dynamic entropy-based triggers in future works.
Moreover, we employ a local lookahead strategy rather than a global search mechanism like MCTS. Although this limits long-horizon planning in extremely complex scenarios, it represents a deliberate trade-off to simultaneously optimize correctness and conciseness, thereby achieving efficiency gains without incurring the heavy computational overhead of exhaustive search.

\section*{Acknowledgments}
This work was partially assisted by AI tools during its development. Specifically, Claude Sonnet 4.5 was used to support code implementation, and Gemini 3.0 Pro was used to assist with writing refinement and language polishing. All scientific contributions, experimental designs, and intellectual content remain solely the work of the authors.

\bibliography{custom}

\appendix

\clearpage

\etocdepthtag{appendix}
\etocsettagdepth{main}{none}
\etocsettagdepth{appendix}{subsubsection} 

\newcommand{\tocdot}{\nobreak\leaders\hbox to 0.8em{\hss.\hss}\hfill}

\etocsetstyle{section}
    {\par}                
    {\addvspace{0.35em}}    
    {\noindent \textbf{\etocnumber{} \etocname} \tocdot \etocpage \par}
    {}                

\etocsetstyle{subsection}
    {} 
    {\addvspace{0.15em}}    
    {\noindent \hspace{0.5em} \etocnumber{} \etocname \tocdot \etocpage \par} 
    {} 

\etocsetstyle{subsubsection}
    {} 
    {} 
    {\noindent \hspace{1.5em} \etocnumber{} \etocname \tocdot \etocpage \par} 
    {}

{
  % \centerline{\large \textbf{Appendix}} 
  % \vspace{0.25em} 
  \setlength{\parskip}{2.25pt}
  \tableofcontents
}

\section{Implementation Details}
\label{appendix:imple}

\subsection{Details of the hybrid guidance experiment}
\label{appendix:hybird_guidance}

\begin{figure}[h]
  \centering
\includegraphics[width=0.98\linewidth]{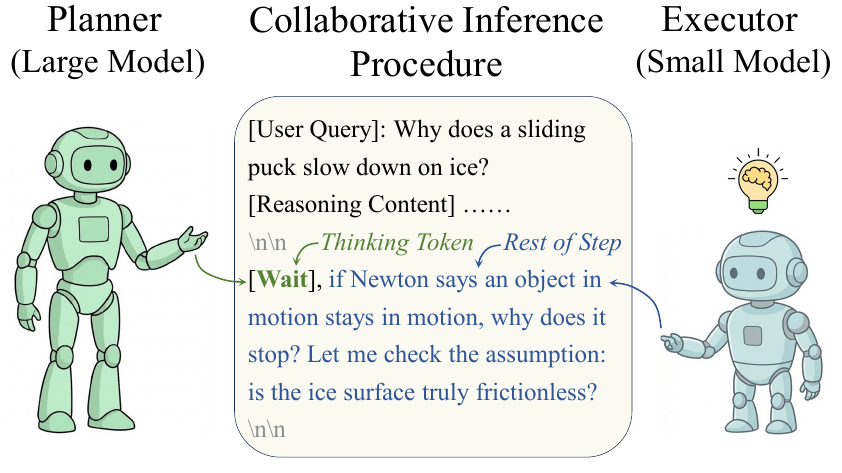}
\vspace{-5pt}
  \caption{Illustration of the Collaborative Inference Procedure. At each reasoning step delimiter (\textbackslash n\textbackslash n), the larger Planner model intervenes to generate a single, strategic Thinking Token (e.g., [Wait]). This token directs the reasoning path, while the smaller Executor model generates the detailed remainder of the step.}
  \vspace{-2pt}
\label{fig:co-gen}
\end{figure}

To empirically validate the hypothesis that Small Reasoning Models (SRMs) suffer primarily from a lack of high-level planning rather than token-level execution, we designed a Hybrid Guidance framework. This framework decouples strategic planning from detailed execution by employing a collaborative generation process between a stronger Planner model and a weaker Executor model.

\subsubsection{Experimental Setup}
We utilize DeepSeek-R1-Distill-Qwen-32B as the strategic planner ($\mathcal{M}_{\text{plan}}$) and DeepSeek-R1-Distill-Qwen-7B as the executor ($\mathcal{M}_{\text{exec}}$). The reasoning process is treated as a sequence of discrete steps, delimited by the token sequence \verb|"\n\n"|.

\subsubsection{Collaborative Inference Procedure}
As shown in Fig.\ref{fig:co-gen}, the generation follows an iterative handover mechanism. The process begins with the input query $x$. At the start of each reasoning step (after a \verb|"\n\n"| delimiter, as introduced in Section \ref{sec:preliminary}), the context is passed to the planner $\mathcal{M}_{\text{plan}}$. The planner is constrained to generate exactly one single token. This token serves as the directional guide (e.g., a logical connective or a reflective marker). Once this guiding token is generated, it is appended to the context. Control is then transferred to the executor $\mathcal{M}_{\text{exec}}$, which generates the remainder of the reasoning step until it predicts the next step delimiter \verb|"\n\n"|. This cycle repeats until the final answer is derived or the maximum token length is reached. 

\paragraph{Performance.} Despite the minimal intervention, where guiding tokens account for only \textbf{2.9\%} of the total generated tokens, the hybrid approach yields a substantial performance improvement. As shown in Fig.\ref{fig:fig1}), the 7B model achieves an average accuracy gain of \textbf{6.2\%} across benchmarks when guided by the 32B model's guiding tokens. These results suggest that SRMs possess sufficient capability for granular execution but struggle to independently navigate complex reasoning paths.

\subsubsection{Analysis of Guiding Tokens}
To examine whether $\mathcal{M}_{\text{plan}}$ injects factual knowledge or structural guidance, we analyzed the frequency distribution of the tokens generated by $\mathcal{M}_{\text{plan}}$ across the test set. As illustrated in Table \ref{tab:guidance_tokens}, the vast majority of generated tokens are logical connectives (defined as Thinking Tokens and Reasoning Operators in our method), such as "Wait", "So", and "Alternatively". Content-heavy nouns or entities are rarely generated during this phase. This confirms that the larger model primarily contributes to the reasoning process by steering the logical flow and correcting the reasoning path rather than providing direct factual answers.

\begin{table}[t]
    \centering
    \small
    \begin{tabular}{lr@{\hspace{2em}}lr}
        \toprule
        \textbf{Token} & \textbf{Freq (\%)} & \textbf{Token} & \textbf{Freq (\%)} \\
        \midrule
        So & 12.15 & I & 1.98 \\
        Wait & 8.75 & Hmm & 1.53 \\
        Let & 5.68 & Then & 1.13 \\
        But & 4.89 & Thus & 0.85 \\
        Option & 4.36 & If & 0.57 \\
        First & 3.54 & Since & 0.53 \\
        Therefore & 3.27 & Given & 0.48 \\
        Now & 3.06 & Simplify & 0.46 \\
        Alternatively & 2.82 & Alright & 0.41 \\
        The & 2.37 & Yes & 0.40 \\
        \bottomrule
    \end{tabular}
    \caption{Frequency Distribution of Guiding Tokens generated by the Planner $\mathcal{M}_{\text{plan}}$. The distribution is heavily dominated by logical connectives (e.g., "So", "Wait"), demonstrating that the Planner provides structural guidance to navigate the reasoning path rather than specific content.}
    \label{tab:guidance_tokens}
    \vspace{-5pt}
\end{table}

\subsection{Details of the random search experiment}
\label{appendix:random_search}

To investigate the feasibility and potential of our proposed search-based framework, we designed a randomized search experiment. The primary objective of this experiment is to probe the boundaries of the solution space and determine whether there exist superior paths, which are defined as reasoning paths that achieve higher accuracy and lower computational cost than those produced by the model's standard generation policy.

\subsubsection{Randomized Intervention Procedure}
We utilize a stochastic intervention mechanism to explore diverse reasoning paths.  Let $\mathcal{M}$ denote the language model and $x$ the input query. During the generation process, we monitor the stream for the step delimiter token sequence \verb|"\n\n"|, which marks the completion of a reasoning step $s_{t-1}$. At this decision point, we suspend the standard sampling process. Instead of selecting the subsequent token from the model's predicted distribution, we uniformly sample a reasoning operator $o_t$ from a fixed set $\mathcal{O}$. Specifically, We define this set as $\mathcal{O} = \{\text{"The"}, \text{"Thus"}, \text{"Therefore"}, \text{"So"}, \text{"Then"},$ $\text{"Let"}, \text{"Wait"}, \text{"Alternatively"}\}$. This operator is forced into the context context as the prefix for step $s_t$. The model $\mathcal{M}$ then resumes generation conditioned on this intervention. This cycle repeats until the model outputs a final answer or reaches a horizon of $T_{max}=50$ steps.

\subsubsection{Construction of the Solution Space}
We characterize the solution space through high-volume sampling. For each query $x_i$ in the evaluation dataset of size $N$, we generate $K=16$ independent reasoning paths via the intervention procedure described above. We visualize the resulting performance distribution (Fig. \ref{fig:distribution}) using Monte Carlo aggregation. A single data point in the density heatmap corresponds to a coordinate pair $(\bar{L}, \bar{A})$, representing the average length and average accuracy over the full dataset. To generate one such point, we traverse all $N$ queries and randomly select exactly one path from the $K$ available candidates for each query. We then calculate the mean length $\bar{L}$ and mean accuracy $\bar{A}$ for this specific combination of selected paths. By repeating this sampling process for a large number of iterations (e.g., $10^6$ times), we obtain a dense distribution of coordinates. This distribution effectively estimates the probability density of the model's performance across the entire feasible solution space.
The region in the solution space where paths exhibit both higher accuracy and lower length than the original baseline confirms the existence of the Superior Paths and validates the motivation for our Neural Chain-of-Thought Search.

\subsection{Details of the Benchmarks considered}
\label{appendix:benchmarks}
We selected five benchmarks to empirically encompass the spectrum of reasoning capabilities: symbolic deductive reasoning, commonsense reasoning, expert knowledge reasoning, multi-step arithmetic reasoning and Olympiad-level mathematical reasoning. This diversity ensures that our observed efficiency gains are substantive and extend beyond any single problem type.

\paragraph{AMC23.} Derived from the 2023 American Mathematics Competitions, this dataset represents a significant step up in difficulty compared to standard arithmetic benchmarks. Unlike grade-school problems, AMC23 requires rigorous multi-step logical deduction and the application of complex mathematical theorems. We use this benchmark to test the model's ability to maintain coherent long-chain reasoning without degenerating into circular logic, a common failure mode in harder deductive tasks.

\paragraph{ARC-C \cite{clark2018thinksolvedquestionanswering}.} The Abstraction and Reasoning Challenge (Challenge Set) evaluates a model's ability to infer abstract rules from few-shot examples. While originally a visual grid-based task, we use the text-encoded version to test the capacity to recognize patterns and generalize to unseen problems. This benchmark is relevant for analyzing thinking tokens, as it demands a search process to hypothesize and verify transformation rules, distinguishing it from pure retrieval tasks.

\paragraph{GPQA \cite{rein2023gpqagraduatelevelgoogleproofqa}.} The Graduate-Level Google-Proof Q\&A benchmark consists of difficult multiple-choice questions in biology, physics, and chemistry. Validated by domain experts who hold or are pursuing PhDs, these questions are designed to be resistant to simple web search. We include GPQA to evaluate the "knowledge-intensive" reasoning regime. Here, the efficiency bottleneck is often not the length of the deduction, but the accuracy of the fact retrieval and the avoidance of "hallucinated reasoning," where models generate verbose justifications for incorrect premises.

\paragraph{GSM8K \cite{cobbe2021trainingverifierssolvemath}.} This widely-used benchmark consists of 8.5k high-quality grade school math word problems that require 2 to 8 steps to solve. While less challenging than AMC23, its arithmetic operations allows us to measure the efficacy of our method in pruning redundant verification steps in well-defined solution spaces.

\paragraph{OlympiadBench \cite{he2024olympiadbench}.} As a comprehensive dataset sourced from international Olympiad-level mathematics and physics competitions, this benchmark presents a formidable challenge to current reasoning models. Unlike the routine application of formulas in GSM8K, OlympiadBench demands creative problem-solving strategies and extended logical derivations. We employ it to evaluate the efficacy of our search mechanism in high complexity regimes, specifically testing its capability to navigate the deep reasoning trees required for creative problem solving.

\begin{figure*}[!ht]
  \centering
  \vspace{-5pt}
\includegraphics[width=0.98\linewidth]{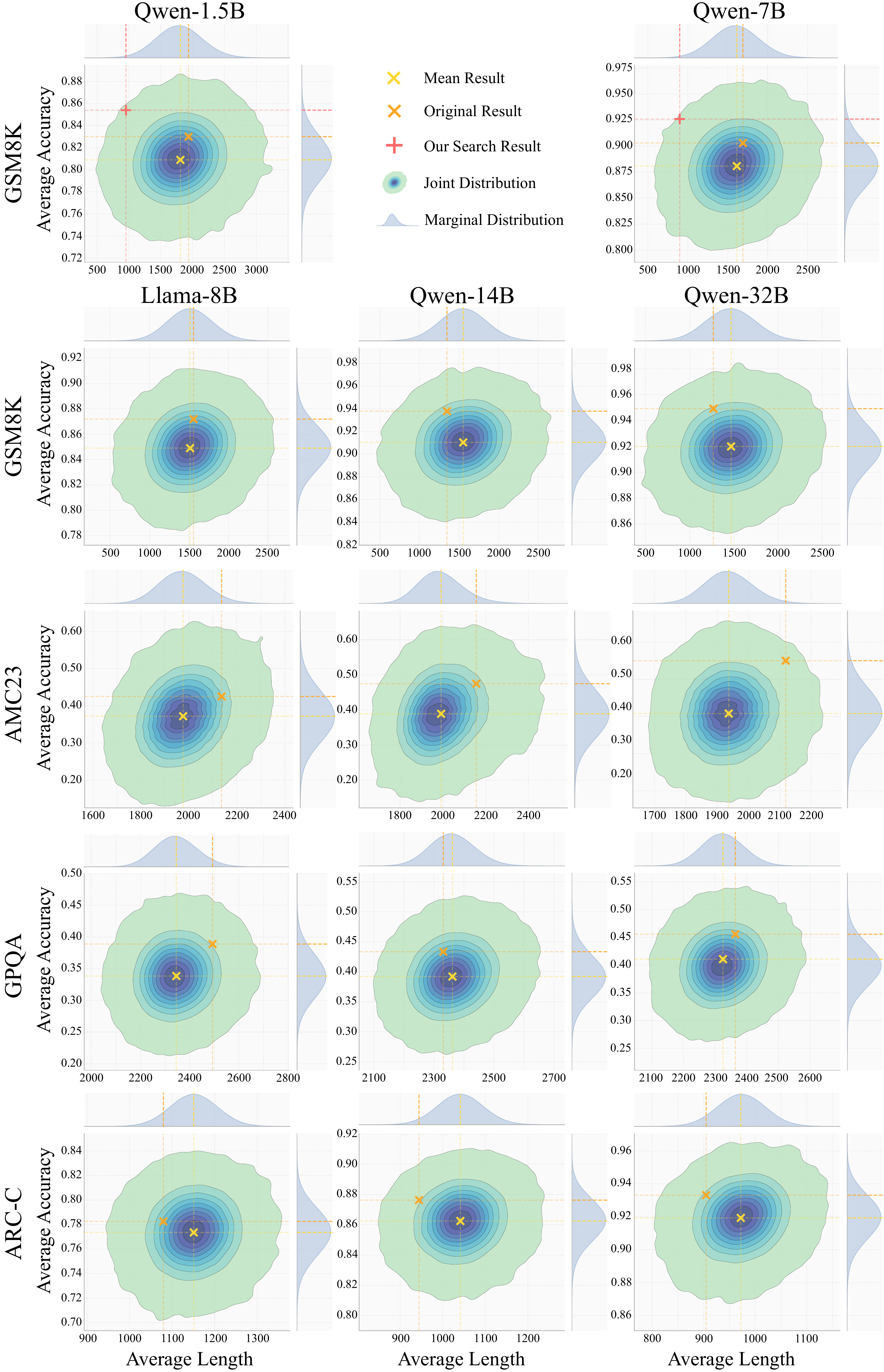}
  \caption{Reasoning solution space visualization across diverse models and benchmarks.}
  \vspace{-5pt}
\label{fig:more_dist}
\end{figure*}

\subsection{Details of the Baselines considered}
\label{appendix:baselines}
\paragraph{Mean and Original.}
The Original baseline represents the standard sampling (temperature=0.6, top-p=0.95) from the base model without intervention. The Mean baseline, as described in Section \ref{sec:solution_space}, represents the expected performance of a random search strategy where operators are sampled uniformly from the set $\mathcal{O}$ at decision points. This comparison isolates the specific contribution of our learned policy network versus a blind search.

\paragraph{NoWait \citep{wang2025waitdontneedwait}.} This training-free decoding strategy operates on the hypothesis that reflective tokens often signal hesitation or redundant loops. NoWait explicitly suppresses the generation of self-reflection tokens (e.g., "Wait", "Hmm") during the decoding process. We include this baseline to demonstrate that naive truncation of reasoning paths often degrades accuracy, whereas our method preserves correctness.

\paragraph{AdaptThink \cite{zhang2025adaptthinkreasoningmodelslearn}.} AdaptThink is a reinforcement learning-based approach that focuses on the extensive margin. It trains the model to adaptively select between "thinking" (long-CoT) and "no-thinking" (direct answer) modes based on the estimated difficulty of the input query. Unlike our fine-grained operator search which structures the internal steps of the reasoning chain, AdaptThink makes a binary, high-level decision on whether to engage the reasoning engine at all.

\paragraph{ThinkPrune \cite{hou2025thinkprunepruninglongchainofthought}.} ThinkPrune addresses the efficiency-accuracy trade-off by incorporating a strict token budget into the reward function during RL training. It penalizes generation length linearly or non-linearly to force the model to compress its reasoning. This baseline serves as a direct comparison for our length-penalty reward component, validating whether our Dual-Factor Heuristic Function offers superior control compared to scalar reward shaping alone.

\paragraph{Laser \citep{liu2025learnreasonefficientlyadaptive}.} Length-bAsed StEp Reward shaping (LASER) is a technique that optimizes the trade-off between performance and efficiency using adaptive length-based incentives. It employs a step-function reward scheme that dynamically adjusts the penalty for length based on the current training stage and problem difficulty. We include LASER as a representative of reward-shaping approaches to efficient reasoning.

\section{Additional Experimental Results}
\label{appendix:more_result}
\subsection{Visualizations of solution spaces for more models and more benchmarks}
\paragraph{Experimental Settings.}
We examine the generalizability of the solution space characteristics by extending the random search analysis to a broader experimental settings. We conduct comprehensive experiments across five models featuring varying parameter scales and architectures: DeepSeek-R1-Distill-Qwen-\{1.5B, 7B, 14B, 32B\} and DeepSeek-R1-Distill-Llama-8B. Furthermore, to ensure robustness across different reasoning modalities, we evaluate these models on four distinct benchmarks: AMC23, ARC-C \cite{clark2018thinksolvedquestionanswering}, GPQA \cite{rein2023gpqagraduatelevelgoogleproofqa}, and GSM8K \cite{cobbe2021trainingverifierssolvemath}. Fig \ref{fig:more_dist} illustrates the density heatmaps for all combinations of models and datasets, generated using the Monte Carlo aggregation method detailed in Appendix \ref{appendix:random_search}.

\paragraph{Consistency of Insights.}
The comprehensive evaluations consistently demonstrate the four fundamental insights discussed in Section \ref{sec:solution_space}: (1) The choice of reasoning operators induces high variance in output quality. (2) Standard decoding strategies consistently result in suboptimal paths relative to the potential maximum. (3) Superior paths that are simultaneous more accurate and efficient than the original output exist across all models and tasks. (4) These superior paths are distributed sparsely within the solution space. 

\paragraph{Impact of Model Scale.}
Beyond these confirmations, we observe a inverse correlation between model scale and the density of improved solutions. Comparing heatmaps reveals that the solution space area superior to the original baseline contracts as model size increases. For instance, on the GSM8K benchmark, the density of superior paths for the $1.5$B model is $9.12\%$, whereas this value drops to $1.30\%$ for the $32$B model. This phenomenon suggests that larger models possess stronger intrinsic planning capabilities. Their default generation policies align more closely with optimal reasoning paths, narrowing the margin for improvement accessible through random exploration.

\section{Related Work}
\label{appendix:related_work}
\subsection{Efficient Reasoning}
Efficient reasoning has emerged as a critical research direction to mitigate the computational overhead and overthinking phenomenon \cite{sui2025stopoverthinkingsurveyefficient, kong2026tokenreductionefficiencygenerative} observed in Large Reasoning Models (LRMs) like DeepSeek-R1 \cite{deepseekai2025deepseekr1incentivizingreasoningcapability} and OpenAI o1 \cite{openai2024openaio1card}. We categorize existing efficient reasoning approaches into four main paradigms: Reinforcement Learning (RL) with length reward design, Supervised Fine-Tuning (SFT) with variable-length data, dynamic reasoning paradigms during inference, and prompt-guided efficiency.

\subsubsection{RL with Length Reward Design}
Reinforcement learning has been widely adopted to enhance reasoning capabilities, yet standard accuracy-based rewards often lead to verbose chains of thought. To address this, recent works incorporate length-based penalties directly into the reward function to encourage conciseness without sacrificing performance. Kimi k1.5 \cite{kimiteam2025kimik15scalingreinforcement} integrates a length penalty into its policy optimization (a variant of online policy mirror descent) to facilitate effective model merging and control long CoT activations. 01-Pruner \cite{luo2025o1prunerlengthharmonizingfinetuningo1like} introduces a Length-Harmonizing Reward combined with a PPO-style loss, optimizing the ratio of CoT lengths between a reference model and the student to shorten reasoning while maintaining accuracy constraints. Similarly, L1 \cite{aggarwal2025l1controllinglongreasoning} modifies training data with length constraints (e.g., "Think for N tokens") before applying policy optimization. Demystifying Long CoT \cite{yeo2025demystifyinglongchainofthoughtreasoning} proposes a Cosine Reward based on a Dirichlet function and an exceed length penalty to stabilize performance and control length growth during RL. DAST \cite{shen2025dastdifficultyadaptiveslowthinkinglarge} employs SimPO \cite{meng2024simposimplepreferenceoptimization} with a constructed length-preference dataset based on a token-length budget, while Arora et al. \cite{arora2025traininglanguagemodelsreason} utilize length-based rewards conditioned on correctness, assigning higher scores to shorter, correct answers. AttnPO \cite{nie2026attnpoattentionguidedprocesssupervision} further exploits the model's own attention heads to provide process-level supervision, distinguishing essential from redundant reasoning steps without additional overhead.
Given the rapid expansion of research in this direction, we summarize other significant contributions in Table \ref{tab:rl_conference_papers} and Table \ref{tab:rl_arxiv_papers}.

\begin{table}[t]
\centering
\small
\begin{tabular}{p{5cm}cc}
\toprule
\textbf{Short Name} & \textbf{Venue} & \textbf{Year} \\
\midrule
Demystifying Long \cite{yeo2025demystifyinglongchainofthoughtreasoning} & ICML & 2025 \\
ASRR \cite{zhang2025continuethinkingadaptivethinking} & EMNLP & 2025 \\
ConciseRL \cite{dumitru2025conciserlconcisenessguidedreinforcementlearning} & EMNLP & 2025 \\
AdaptThink \cite{zhang2025adaptthinkreasoningmodelslearn} & EMNLP & 2025 \\
BRPO \cite{qi2025optimizinganytimereasoningbudget} & NeurIPS & 2025 \\
ACPO \cite{cheng2025incentivizingdualprocessthinking} & NeurIPS & 2025 \\
HGPO \cite{jiang2025thinkneedlargehybridreasoning} & NeurIPS & 2025 \\
S-GRPO \cite{dai2025sgrpoearlyexitreinforcement} & NeurIPS & 2025 \\
DeGRPO \cite{fang2025thinklessllmlearnsthink} & NeurIPS & 2025 \\
REO-RL \cite{gao2025faroptimalreasoningefficiency} & NeurIPS & 2025 \\
LIMOPro \cite{xiao2025limoproreasoningrefinementefficient} & NeurIPS & 2025 \\
AutoThink \cite{tu2025learningthinkshapingadaptive} & NeurIPS & 2025 \\
Arora et al. \cite{arora2025traininglanguagemodelsreason} & NeurIPS & 2025 \\
80/20 rule \cite{wang20258020rulehighentropyminority} & NeurIPS & 2025 \\
\bottomrule
\end{tabular}
\caption{Summary of peer-reviewed Conference Papers addressing Efficient Reasoning through RL with Length Reward Design.}
\label{tab:rl_conference_papers}
\end{table}

\begin{table}[t]
\centering
\small
\begin{tabular}{p{5cm}cc}
\toprule
\textbf{Short Name} & \textbf{Venue} & \textbf{Year} \\
\midrule
L1 \cite{aggarwal2025l1controllinglongreasoning} & arXiv & 2025 \\
MRT \cite{qu2025optimizingtesttimecomputemeta} & arXiv & 2025 \\
DTO \cite{an2025dontthinklongerthink} & arXiv & 2025 \\
ALP \cite{xiang2025justthinkingefficientreasoning} & arXiv & 2025 \\
PLP \cite{ling2025fasteasydeephard} & arXiv & 2025 \\
FCS \cite{hong2025reconsideringoverthinkingpenalizinginternal} & arXiv & 2025 \\
DAST \cite{shen2025dastdifficultyadaptiveslowthinkinglarge} & arXiv & 2025 \\
AALC \cite{li2025aalclargelanguagemodel} & arXiv & 2025 \\
SCPO \cite{he2025smartthinkerlearningcompresspreserve} & arXiv & 2025 \\
GFPO \cite{shrivastava2025samplethinklessgroup} & arXiv & 2025 \\
VSRM \cite{yue2025promotingefficientreasoningverifiable} & arXiv & 2025 \\
Bingo \cite{liu2025bingoboostingefficientreasoning} & arXiv & 2025 \\
TL;DR \cite{li2025tldrlongreweightingefficient} & arXiv & 2025 \\
LASER \cite{liu2025learnreasonefficientlyadaptive} & arXiv & 2025 \\
LC-RL \cite{cheng2025optimizinglengthcompressionlarge} & arXiv & 2025 \\
SABER \cite{zhao2025saberswitchablebalancedtraining} & arXiv & 2025 \\
L-GRPO \cite{song2025walkrunconcisellm} & arXiv & 2025 \\
GRPO-X \cite{dai2025stablereinforcementlearningefficient} & arXiv & 2025 \\
DuP-PO \cite{ding2025thinkingtokenshelptrap} & arXiv & 2025 \\
DR.SAF \cite{chen2025awarefirstthinkless} & arXiv & 2025 \\
AdaCoT \cite{lou2025adacotparetooptimaladaptivechainofthought} & arXiv & 2025 \\
HAWKEYE \cite{she2025hawkeyeefficientreasoningmodelcollaboration} & arXiv & 2025 \\
Short-RL \cite{yuan2025efficientrltrainingreasoning} & arXiv & 2025 \\
01-Pruner \cite{luo2025o1prunerlengthharmonizingfinetuningo1like} & arXiv & 2025 \\
LongShort \cite{ning2025thoughtsgeneratedequalefficient} & arXiv & 2025 \\
ThinkPrune \cite{hou2025thinkprunepruninglongchainofthought} & arXiv & 2025 \\
SelfBudgeter \cite{li2025selfbudgeteradaptivetokenallocation} & arXiv & 2025 \\
Self-adaptive \cite{yang2025thinkneedselfadaptivechainofthought} & arXiv & 2025 \\
Concise Reasoning \cite{fatemi2025concisereasoningreinforcementlearning} & arXiv & 2025 \\
Elastic Reasoning \cite{xu2025scalablechainthoughtselastic} & arXiv & 2025 \\
CurriculumGRPO \cite{hammoud2025trainlongthinkshort} & arXiv & 2025 \\
\bottomrule
\end{tabular}
\caption{Summary of recent arXiv Preprints addressing Efficient Reasoning through RL with Length Reward Design.}
\vspace{-5pt}
\label{tab:rl_arxiv_papers}
\end{table}

\subsubsection{SFT with Variable-Length CoT Data}
Fine-tuning LLMs on curated variable-length CoT datasets is another effective strategy to distill efficient reasoning capabilities. These methods generally fall into two categories: post-reasoning compression and during-reasoning compression. In post-reasoning compression, Distilling System 2 into System 1 \cite{yu2024distilling21} removes the reasoning process entirely to distill direct answer generation. C3oT \cite{kang2024c3otgeneratingshorterchainofthought} utilizes GPT-4 as a compressor to reduce reasoning length while retaining key information. TokenSkip \cite{xia2025tokenskipcontrollablechainofthoughtcompression} reduces tokens based on semantic importance estimation. In during-reasoning compression, Learn to Skip \cite{liu2024languagemodelslearnskip} adopts a human-like step-skipping method, first manually creating concise solutions and then training the model to intrinsically skip steps. Token-Budget \cite{han2025tokenbudgetawarellmreasoning} employs a binary search to find optimal token budgets and trains the model to follow these constraints. Self-Training \cite{munkhbat2025selftrainingelicitsconcisereasoning} uses Best-of-N sampling to select the shortest correct reasoning path as training data. CoT-Valve \cite{ma2025cotvalvelengthcompressiblechainofthoughttuning} progressively mixes parameters of long-reasoning and non-reasoning models to generate variable-length training data. To provide a structured overview of the rapidly evolving landscape, we summarize relevant peer-reviewed conference papers in Table \ref{tab:sft_conference_papers} and recent arXiv preprints in Table \ref{tab:sft_arxiv_papers}.

\begin{table}[h]
\centering
\small
\begin{tabular}{p{5cm}cc}
\toprule
\textbf{Short Name} & \textbf{Venue} & \textbf{Year} \\
\midrule
Stepwise \cite{cui2025stepwiseperplexityguidedrefinementefficient} & ACL & 2025 \\
CoT-Valve \cite{ma2025cotvalvelengthcompressiblechainofthoughttuning} & ACL & 2025 \\
Token-Budget \cite{han2025tokenbudgetawarellmreasoning} & ACL & 2025 \\
Self-Training \cite{munkhbat2025selftrainingelicitsconcisereasoning} & ACL & 2025 \\
C3oT \cite{kang2024c3otgeneratingshorterchainofthought} & AAAI & 2025 \\
ReCUT \cite{jin2025recutbalancingreasoninglength} & EMNLP & 2025 \\
ConCISE \cite{qiao2025conciseconfidenceguidedcompressionstepbystep} & EMNLP & 2025 \\
TokenSkip \cite{xia2025tokenskipcontrollablechainofthoughtcompression} & EMNLP & 2025 \\
Learn to Skip \cite{liu2024languagemodelslearnskip} & NeurIPS & 2024 \\
PNS \cite{yu2025causalsufficiencynecessityimproves} & NeurIPS & 2025 \\
Ada-R1 \cite{luo2025adar1hybridcotbileveladaptive} & NeurIPS & 2025 \\
A-Thought* \cite{xu2025athoughtefficientreasoningbidirectional} & NeurIPS & 2025 \\
VeriThinker \cite{chen2025verithinkerlearningverifymakes} & NeurIPS & 2025 \\
\bottomrule
\end{tabular}
\caption{Summary of peer-reviewed Conference Papers addressing Efficient Reasoning through SFT with Variable-Length CoT Data.}
\vspace{-5pt}
\label{tab:sft_conference_papers}
\end{table}

\begin{table}[h]
\centering
\small
\begin{tabular}{p{5cm}cc}
\toprule
\textbf{Short Name} & \textbf{Venue} & \textbf{Year} \\
\midrule
Distilling System 2 \cite{yu2024distilling21} & arXiv & 2024 \\
Z1 \cite{yu2025z1efficienttesttimescaling} & arXiv & 2025 \\
DRP \cite{jiang2025drpdistilledreasoningpruning} & arXiv & 2025 \\
CTS \cite{yuan2025tokensneedthinking} & arXiv & 2025 \\
ASAP \cite{zeng2025pruningunsurprisingefficientcode} & arXiv & 2025 \\
AutoL2S \cite{luo2025autol2sautolongshortreasoning} & arXiv & 2025 \\
Verbosity \cite{jang2025verbosityawarerationalereductioneffective} & arXiv & 2025 \\
OThink-R1 \cite{zhang2025othinkr1intrinsicfastslowthinking} & arXiv & 2025 \\
R1-Compress \cite{wang2025r1compresslongchainofthoughtcompression} & arXiv & 2025 \\
StepEntropy \cite{li2025compressingchainofthoughtllmsstep} & arXiv & 2025 \\
Prune-on-Logic \cite{zhao2025pruningimprovereasoningrevisiting} & arXiv & 2025 \\
LS-Mixture SFT \cite{yu2025longshortchainofthoughtmixturesupervised} & arXiv & 2025 \\
Assembly of Experts \cite{klagges2025assemblyexpertslineartimeconstruction} & arXiv & 2025 \\
\bottomrule
\end{tabular}
\caption{Summary of recent arXiv Preprints addressing Efficient Reasoning through SFT with Variable-Length CoT Data.}
\vspace{-5pt}
\label{tab:sft_arxiv_papers}
\end{table}

\subsubsection{Inference Time Dynamic Reasoning}
Dynamic reasoning aims to optimize the inference process without extensive retraining, often by selecting efficient reasoning paths or terminating early. Reward-Guided Efficient Reasoning: Speculative Rejection \cite{sun2024fastbestofndecodingspeculative} optimizes Best-of-N decoding by using a reward model to periodically reject unpromising sequences, reducing computational overhead. RSD \cite{liao2025rewardguidedspeculativedecodingefficient} employs a Process Reward Model (PRM) to selectively accept high-quality outputs from a draft model. Confidence and Certainty-Based Adaptive Reasoning: DPTS \cite{ding2025dynamicparalleltreesearch} optimizes tree search by dynamically adjusting node expansion based on confidence. FastMCTS \cite{li2025fastmctssimplesamplingstrategy} prioritizes high-confidence traces in an MCTS-inspired framework. Certaindex \cite{fu2025efficientlyscalingllmreasoning} and Dynasor \cite{fu2025reasoning} allocate compute based on a statistical measure of reasoning progress. Length-filtered Vote \cite{wu2025lessunderstandingchainofthoughtlength} filters out excessively short or long paths before majority voting. CISC \cite{Taubenfeld_2025} utilizes confidence scores to implement early stopping in sampling. Consistency-Based Reasoning: ST-BoN \cite{wang2025samplingefficienttesttimescalingselfestimating} leverages the consistency of latent embeddings to truncate inferior samples early, serving as a proxy for answer correctness. Summarization-Based Reasoning: LightThinker \cite{zhang2025lightthinkerthinkingstepbystepcompression} trains models to compress intermediate thoughts into "gist tokens," while InftyThink \cite{yan2025inftythinkbreakinglengthlimits} iteratively summarizes thoughts to enable unbounded reasoning depth within context limits. Given the rapid proliferation of strategies in this field, we provide a comprehensive summary of other significant contributions, categorized into peer-reviewed conference papers and recent preprints, in Table \ref{tab:inf_conference_papers} and Table \ref{tab:inf_arxiv_papers}, respectively.

\begin{table}[t]
\centering
\small
\begin{tabular}{p{5cm}cc}
\toprule
\textbf{Short Name} & \textbf{Venue} & \textbf{Year} \\
\midrule
DPTS \cite{ding2025dynamicparalleltreesearch} & ACL & 2025 \\
CISC \cite{Taubenfeld_2025} & ACL & 2025 \\
FastMCTS \cite{li2025fastmctssimplesamplingstrategy} & ACL & 2025 \\
ESC \cite{li2024escapeskyhighcostearlystopping} & ICLR & 2024 \\
RSD \cite{liao2025rewardguidedspeculativedecodingefficient} & ICML & 2025 \\
SpecSearch \cite{wang2025acceleratinglargelanguagemodel} & ICML & 2025 \\
Best-Route \cite{ding2025bestrouteadaptivellmrouting} & ICML & 2025 \\
AdaptiveStep \cite{liu2025adaptivestepautomaticallydividingreasoning} & ICML & 2025 \\
Adaptive Reasoning \cite{yu2025thinksmarterharderadaptive} & ICML & 2025 \\
DSC \cite{wang2025makepennycountdifficultyadaptive} & NAACL & 2025 \\
RASC \cite{wan2025reasoningawareselfconsistencyleveraging} & NAACL & 2025 \\
STAND \cite{Song_2025} & EMNLP & 2025 \\
AlphaOne \cite{zhang2025alphaonereasoningmodelsthinking} & EMNLP & 2025 \\
LightThinker \cite{zhang2025lightthinkerthinkingstepbystepcompression} & EMNLP & 2025 \\
Answer Convergence \cite{liu2025answerconvergencesignalearly} & EMNLP & 2025 \\
Dynasor \cite{fu2025reasoning} & ICLR'WS & 2025 \\
Speculative Rejection \cite{sun2024fastbestofndecodingspeculative} & NeurIPS & 2024 \\
VGS \cite{wang2025valueguidedsearchefficientchainofthought} & NeurIPS & 2025 \\
RPC \cite{song2025reasoningpathcompressioncompressing} & NeurIPS & 2025 \\
DORA \cite{wang2025rolloutcountsoptimalresource} & NeurIPS & 2025 \\
TOPS \cite{yang2025thinkingoptimalscalingtesttimecompute} & NeurIPS & 2025 \\
ST-BoN \cite{wang2025samplingefficienttesttimescalingselfestimating} & NeurIPS & 2025 \\
ThinkLess \cite{fang2025thinklessllmlearnsthink} & NeurIPS & 2025 \\
SpecReason \cite{pan2025specreasonfastaccurateinferencetime} & NeurIPS & 2025 \\
\bottomrule
\end{tabular}
\caption{Summary of peer-reviewed Conference Papers addressing Efficient Reasoning through Inference Time Dynamic Reasoning. WS denotes workshop.}
\vspace{-5pt}
\label{tab:inf_conference_papers}
\end{table}

\begin{table}[h]
\centering
\small
\begin{tabular}{p{5cm}cc}
\toprule
\textbf{Short Name} & \textbf{Venue} & \textbf{Year} \\
\midrule
GG \cite{ghasemabadi2025guidedgutefficienttesttime} & arXiv & 2025 \\
DO \cite{hassid2025dontoverthinkitpreferring} & arXiv & 2025 \\
FFS \cite{agarwal2025finishsearchefficienttesttime} & arXiv & 2025 \\
CAR \cite{lu2025prolongedreasoningneedcertaintybased} & arXiv & 2025 \\
ASC \cite{azizi2025activationsteeringchainofthoughtcompression} & arXiv & 2025 \\
MUR \cite{yan2025murmomentumuncertaintyguided} & arXiv & 2025 \\
SCoT \cite{wang2025efficientreasoningllmsspeculative} & arXiv & 2025 \\
GoGl \cite{zhuang2025acceleratingchainofthoughtreasoninggoalgradient} & arXiv & 2025 \\
TTPI \cite{yang2025testtimepromptintervention} & arXiv & 2025 \\
CGRS \cite{huang2025efficientreasoninglargereasoning} & arXiv & 2025 \\
SPECS \cite{cemri2025textttspecsfastertesttimescaling} & arXiv & 2025 \\
PathC \cite{zhu2025pathconsistencyprefixenhancementefficient} & arXiv & 2025 \\
TrimR \cite{lin2025trimrverifierbasedtrainingfreethinking} & arXiv & 2025 \\
NOWAIT \cite{wang2025waitdontneedwait} & arXiv & 2025 \\
FracCoT \cite{liao2025fracturedchainofthoughtreasoning} & arXiv & 2025 \\
CoThink \cite{fan2025pricesecondthoughtevaluation} & arXiv & 2025 \\
R-Stitch \cite{chen2025rstitchdynamictrajectorystitching} & arXiv & 2025 \\
ValueFree \cite{sareen2025puttingvaluerlbetter} & arXiv & 2025 \\
Certaindex \cite{fu2025efficientlyscalingllmreasoning} & arXiv & 2025 \\
InftyThink \cite{yan2025inftythinkbreakinglengthlimits} & arXiv & 2025 \\
FlashThink \cite{jiang2025flashthinkearlyexitmethod} & arXiv & 2025 \\
NoThinking \cite{ma2025reasoningmodelseffectivethinking} & arXiv & 2025 \\
Self-Guided \cite{zhao2025exploringexploitinginherentefficiency} & arXiv & 2025 \\
ThoughtMani \cite{liu2025thoughtmanipulationexternalthought} & arXiv & 2025 \\
Retro-Search \cite{lu2025retrosearchexploringuntakenpaths} & arXiv & 2025 \\
ThinkDeepFast \cite{wang2025thinkdeepthinkfast} & arXiv & 2025 \\
Collaborative \cite{lee2025efficientllmcollaborationplanning} & arXiv & 2025 \\
BudgetGuidance \cite{li2025steeringllmthinkingbudget} & arXiv & 2025 \\
Plan and Budget \cite{lin2025planbudgeteffectiveefficient} & arXiv & 2025 \\
Self-Affirmation \cite{liu2025efficientreasoningsuppressionselfaffirmation} & arXiv & 2025 \\
Sleep-time Compute \cite{lin2025sleeptimecomputeinferencescaling} & arXiv & 2025 \\
Length-filtered Vote \cite{wu2025lessunderstandingchainofthoughtlength} & arXiv & 2025 \\
Searching Skeleton \cite{zhang2025searchingmetareasoningskeleton} & arXiv & 2025 \\
Speculative Thinking \cite{yang2025speculativethinkingenhancingsmallmodel} & arXiv & 2025 \\
\bottomrule
\end{tabular}
\caption{Summary of recent arXiv Preprints addressing Efficient Reasoning through Inference Time Dynamic Reasoning.}
\vspace{-5pt}
\label{tab:inf_arxiv_papers}
\end{table}

\subsubsection{Prompt-Guided Efficient Reasoning}
Explicit prompting offers a lightweight mechanism to enforce efficiency. Token-Budget \cite{han2025tokenbudgetawarellmreasoning} (TALE-EP) estimates a minimal token requirement and explicitly prompts the model to adhere to it. Chain of Draft (CoD) \cite{xu2025chaindraftthinkingfaster} encourages the model to write only a minimum draft (e.g., limiting steps to 5 words), finding that this preserves accuracy while reducing verbosity. Token Complexity \cite{lee2025llmscompresschainofthoughttoken} analyzes the trade-off between prompt-based compression and accuracy. Concise CoT (CCoT) \cite{Renze_2024} simply prompts models to "be concise," while MARP \cite{chen2024unlockingcapabilitiesthoughtreasoning} limits single-step computations to refine reasoning boundaries. Other works investigating prompt-based efficiency include Brevity \cite{poddar2025brevitysoulsustainabilitycharacterizing}, PREMISE \cite{yu2025premisescalablestrategicprompt}, GUARD \cite{ding2025guardglocaluncertaintyawarerobust} and ConciseHint \cite{tang2025concisehintboostingefficientreasoning}.

\subsubsection{Related Benchmarks and Evaluations} 
The development of efficient reasoning methods is supported by parallel works on specialized benchmarks. 
 These benchmarks primarily focus on two interconnected aspects: identifying the \textit{overthinking} pathology in LRMs and establishing standardized metrics for assessing the trade-off between reasoning quality and computational cost. For problem diagnosis, benchmarks like \cite{hashemi2025dnrbenchbenchmarkingoverreasoning} and \cite{zhang2025s1} are designed to trigger and measure excessive verbosity on trivial or intuitive tasks, revealing a deep-seated reasoning bias. Similarly, \cite{srivastava2025llmthinkbench} provides fine-grained analysis of overthinking patterns in basic math, while \citet{zhang2026logicalphasetransitionsunderstanding} identify abrupt performance collapse beyond critical complexity thresholds. To evaluate mitigation strategies and model calibration, benchmarks such as \cite{pu2025thoughtterminator} and \cite{li2025think} introduce metrics like token efficiency and CoT precision/recall. Moving towards a holistic evaluation, unified frameworks like \cite{aggarwal2025optimalthinkingbench} and \cite{huang2025effireason} formalize the dual challenge of preventing waste on easy tasks while ensuring sufficient thought for hard ones, using composite scores like the E3-Score. 
 Beyond these, parallel works also encompass benchmarks for evaluating long context understanding \cite{huang2025minilongbench}, routing mechanisms in LLMs \cite{huang2025routereval}, creative thinking \cite{lotbench, zhong2024let}, agentic tasks \cite{ling2026agent}, challenging math competitions \cite{an2025amobenchlargelanguagemodels}, and general reasoning across diverse tasks \cite{liu2026general365benchmarkinggeneralreasoning}.
 These benchmarks collectively provide the ground truth for developing and comparing the RL, SFT, dynamic, and prompt-guided methods discussed in prior subsections. 

\subsubsection{Connection to Our Work}
Distinct from RL \citep{kimiteam2025kimik15scalingreinforcement, luo2025o1prunerlengthharmonizingfinetuningo1like, yeo2025demystifyinglongchainofthoughtreasoning, zhou2026lookinwardexploreoutward} and SFT \citep{kang2024c3otgeneratingshorterchainofthought, ma2025cotvalvelengthcompressiblechainofthoughttuning, xia2025tokenskipcontrollablechainofthoughtcompression, yu2024distilling21} approaches that enforce efficiency via static training objectives, our method avoids inducing a fixed length bias. We instead formulate efficiency as a dynamic search objective. This decoupling enables adaptive compute allocation; the model expands reasoning for complex queries and prunes redundancy for simpler ones. We therefore prevent the performance degradation frequently observed with forced conciseness \citep{li2025smallmodelsstrugglelearn, jin2024impactreasoningsteplength}. Since we intervene at inference time, our strategy remains orthogonal to these training-based optimizations.
Our framework aligns with dynamic reasoning paradigms \citep{sun2024fastbestofndecodingspeculative, ding2025dynamicparalleltreesearch, li2025fastmctssimplesamplingstrategy} utilizing test-time compute, yet introduces a structural shift in the search space. While prior work relies on token-level search \citep{sun2024fastbestofndecodingspeculative} or heuristic early stopping \citep{fu2025efficientlyscalingllmreasoning}, we reformulate CoT generation as a dynamic search over discrete reasoning operators. By abstracting tokens into operators, we resolve the high-level planning bottleneck. This renders the search strategic and computationally feasible compared to unstructured sampling \citep{wang2025samplingefficienttesttimescalingselfestimating}.

\subsection{Test-Time Compute via Search}
% A Survey on LLM Test-Time Compute via Search
% Demystifying Chains, Trees, and Graphs of Thoughts
The paradigm of scaling test-time compute has emerged as a critical frontier in enhancing LLM reasoning. This field can be broadly categorized through two complementary lenses: the topology of reasoning (the structural connection of thoughts) and the search algorithms (the control policies for traversing these structures).

\subsubsection{Structured Reasoning Topologies} This line of research focuses on the structural representation of intermediate reasoning steps, moving from linear sequences to complex non-linear structures \cite{Besta_2025}.

\paragraph{Chain-based Structures} The seminal Chain-of-Thought (CoT) \cite{wei2023chainofthoughtpromptingelicitsreasoning} prompting demonstrated that eliciting intermediate reasoning steps significantly boosts performance on complex tasks. This linear topology models reasoning as a sequential path graph. Extensions such as CoT-SC (Self-Consistency) \cite{wang2023selfconsistencyimproveschainthought} introduce a "Tree of Chains" topology by sampling multiple independent reasoning paths and aggregating the final answer via majority voting. While effective, chain-based methods suffer from error propagation in long-horizon tasks, as they lack mechanisms to explore alternative branches once a step is generated.

\paragraph{Tree-based Structures} To enable exploration and backtracking, Tree of Thoughts (ToT) \cite{yao2023treethoughtsdeliberateproblem} and \cite{long2023largelanguagemodelguided} generalize CoT by modeling reasoning as a tree, where nodes represent partial solutions or "thoughts". This allows the model to explore multiple reasoning branches at each step. Variants such as Thought Decomposition \cite{xie2023selfevaluationguidedbeamsearch} and Tree-of-Mixed-Thought \cite{hu2023treeofmixedthoughtcombiningfastslow} further refine this by varying the granularity of tree nodes. Other works like Skeleton-of-Thought \cite{ning2024skeletonofthoughtpromptingllmsefficient} utilize a parallel tree structure (or 1-level tree) to accelerate generation by expanding independent points simultaneously. While tree topologies allow for local exploration, they often require manually defining the branching factor and depth.

\paragraph{Graph-based Structures} Graph of Thoughts (GoT) \cite{Besta_2024} and \cite{yao-etal-2024-got} further extend reasoning topologies to arbitrary directed acyclic graphs (DAGs). These frameworks introduce aggregation operations, allowing information from multiple independent reasoning paths to be combined into a synergistic solution. Similarly, Cumulative Reasoning \cite{zhang2025cumulativereasoninglargelanguage} and Everything of Thoughts (XoT) \cite{ding2024thoughtsdefyinglawpenrose} utilize graph structures to model complex dependencies where a thought may depend on multiple non-consecutive precursors. While powerful, graph-based methods incur significant computational overhead due to the complexity of managing arbitrary dependencies. Other graph-based reasoning works include Weight-of-Thought \cite{punjwani2025weightofthoughtreasoningexploringneural}, LogicAgent \cite{zhang2026semanticawarelogicalreasoningsemiotic} and StrucSum \cite{yuan2026strucsumgraphstructuredreasoninglong}.

\subsubsection{Search Algorithms and Planning}Parallel to structural definitions, significant research focuses on the algorithmic procedures used to traverse the reasoning space. These approaches typically formulate the reasoning task as an MDP defined by states, actions, and rewards \cite{li2025surveyllmtesttimecompute}.

\paragraph{Uninformed and Heuristic Search}Early attempts applied standard search algorithms to LLM decoding. Beam Search, as utilized in Beam-LLM \cite{xie2023selfevaluationguidedbeamsearch} and PathFinder \cite{golovneva2023pathfinderguidedsearchmultistep}, maintains the top-$k$ most promising partial candidates at each step. To guide the search more effectively, heuristic methods like Best-First Search \cite{koh2025treesearchlanguagemodel} and A* Search have been adapted. For instance, LLM-A* \cite{meng2025llmalargelanguagemodel} and ToolChain* \cite{zhuang2023toolchainefficientactionspace} integrate cost-to-go heuristics (often estimating the distance to the goal) to prioritize the expansion of promising nodes. $Q^*$ \cite{wang2024qimprovingmultistepreasoning} further approximates optimal Q-values to guide the search using A*-like heuristics. These methods rely heavily on the quality of the heuristic function, which is often difficult to define for open-ended reasoning tasks.

\paragraph{Monte Carlo Tree Search (MCTS)}MCTS has become a dominant paradigm for solving complex reasoning tasks due to its ability to balance exploration and exploitation. Frameworks such as RAP \cite{hao2023reasoninglanguagemodelplanning}, LATS \cite{zhou2024languageagenttreesearch}, and LLM-MCTS \cite{zhao2023largelanguagemodelscommonsense} employ MCTS to simulate future outcomes (rollouts) and backpropagate value estimates to the current state. Recent advancements like rStar \cite{qi2024mutualreasoningmakessmaller} and MC-DML \cite{shi2025montecarloplanninglarge} introduce specialized selection policies and self-consistency-based evaluations to enhance MCTS in reasoning domains. Furthermore, AlphaZero-inspired approaches like TS-LLM \cite{feng2024alphazeroliketreesearchguidelarge} and ReST-MCTS* \cite{zhang2024restmctsllmselftrainingprocess} integrate MCTS with model training, using the search results to iteratively fine-tune the policy and value networks.

\subsubsection{Connection to Our Work}While existing search-based inference methods demonstrate strong performance, they typically rely on heavy sampling, expensive rollouts (as in MCTS), or rigid topological constraints (as in ToT). Our proposed Neural Chain-of-Thought Search differs by internalizing the search process. Instead of managing an external search tree, we treat the discrete thinking tokens as the action space in a Neural Architecture Search (NAS) formulation. This allows our model to dynamically learn a lightweight policy that steers the reasoning topology on-the-fly, achieving the benefits of structured search with significantly lower inference latency than traditional MCTS or massive parallel sampling.

\subsection{AutoML}
Automated Machine Learning (AutoML) aims to automate the end-to-end process of applying machine learning to real-world problems, thereby reducing the reliance on human expertise and manual trial-and-error \cite{He_2021}. The scope of AutoML is broad, covering various stages of the deep learning pipeline including data preparation, feature engineering, hyperparameter optimization (HPO), and model generation \cite{shen2024automatedmachinelearningprinciples}. In the realm of data preparation and feature engineering, techniques have been developed for automated data cleaning, synthesis, and feature selection to maximize the predictive power of raw data \cite{Chu2016DataCO}. However, the most computationally intensive aspect of AutoML lies in model selection and optimization. Traditional approaches focused heavily on Hyperparameter Optimization (HPO) to tune static parameters such as learning rates or batch sizes using methods like Grid Search, Random Search \cite{JMLR:v13:bergstra12a}, Bayesian Optimization \cite{snoek2012practicalbayesianoptimizationmachine}, and bandit-based strategies like Hyperband \cite{li2018hyperbandnovelbanditbasedapproach}. With the advent of deep learning, the focus of AutoML has progressively shifted from tuning hyperparameters of fixed models to the automatic discovery of the model structure itself, leading to the emergence of Neural Architecture Search (NAS).

\subsubsection{Neural Architecture Search} Neural Architecture Search (NAS) is a prominent subfield of AutoML dedicated to automating the design of neural network topologies, which has successfully produced architectures surpassing manually designed counterparts in tasks like image classification and object detection \cite{elsken2019neuralarchitecturesearchsurvey}. A standard NAS framework is typically categorized into three dimensions: search space, search strategy, and performance estimation strategy \cite{elsken2019neuralarchitecturesearchsurvey}. The search space defines the set of representable architectures, evolving from simple chain-structured sequences to complex cell-based search spaces \cite{zoph2018learningtransferablearchitecturesscalable} and hierarchical representations \cite{liu2018hierarchicalrepresentationsefficientarchitecture}. Regarding search strategies, early works utilized Reinforcement Learning, where a controller RNN samples architectures and is trained via policy gradient to maximize validation accuracy \cite{zoph2017neuralarchitecturesearchreinforcement}. Evolutionary Algorithms (EA) have also proven effective by evolving a population of architectures through mutation and crossover operations \cite{real2019regularizedevolutionimageclassifier}. To mitigate the prohibitive computational costs of training each candidate from scratch, recent research has pivoted towards efficiency. This includes differentiable search methods like DARTS \cite{liu2019dartsdifferentiablearchitecturesearch} which relax the discrete search space to allow gradient-based optimization, and One-Shot methods \cite{pmlr-v80-bender18a} that utilize weight sharing within a supernet \cite{pham2018efficientneuralarchitecturesearch}. Furthermore, resource-aware NAS has gained traction, where objective functions are modified to penalize computational costs such as FLOPs or latency \cite{tan2019mnasnetplatformawareneuralarchitecture}, explicitly balancing performance with efficiency \cite{cai2019proxylessnasdirectneuralarchitecture}.
Similar principles of architecture optimization and automatic design have been explored in diverse domains, including diffusion models \cite{huang2023scalelong, zhong2023adapter, lin2024mirrordiffusion}, convolutional network pruning \cite{huang2021rethinking}, and specialized vision tasks \cite{shi2024diff, lu2024sirst}.

\begin{figure*}[h]
\centering
\begin{tcolorbox}[colback=white,colframe=myBlue,title=Prompt1, coltitle=black]
You are an expert reasoning analyst.

The following is one reasoning step extracted from a long chain-of-thought.

Classify its thinking pattern into exactly one of:

- "statement" (presenting facts, calculations, or direct reasoning)

- "reflection" (self-check, doubt, or correction)

- "summary" (concluding or asserting finality)

- "divergence" (exploring alternative paths or possibilities)

Step: """\{Thus, the number of digits is \(\boxed{18}\). \}"""

Return only: \{"label": "<one of statement | reflection | summary | divergence>"\}
\end{tcolorbox}
\caption{The definition-based prompt template for classifying thinking modes based on static definitions.}
\label{appendix:prompt1}
\end{figure*}

\begin{figure*}[h]
\centering
\begin{tcolorbox}[colback=white, colframe=myGreen,title=Prompt2, coltitle=black]
You are an expert reasoning analyst.

The following is one reasoning step extracted from a long chain-of-thought.

Analyze the functional role of this step within the problem-solving flow and classify it into one of:

- "statement" (advancing the flow by deriving new information, calculating, or deducing next steps)

- "reflection" (pausing the flow to critique, verify, or validate the correctness of previous steps)

- "summary" (closing the flow by synthesizing results or explicitly stating the final conclusion)

- "divergence" (branching the flow, shifting strategy, or proposing alternative hypotheses)

Step: """\{Thus, the number of digits is \(\boxed{18}\). \}"""

Return only: \{"label": "<one of statement | reflection | summary | divergence>"\}
\end{tcolorbox}
\caption{The function-based prompt template for analyzing the role of reasoning steps within the reasoning flow.}
\label{appendix:prompt2}
\end{figure*}

\subsubsection{Connection to Our Work} Our proposed Neural Chain-of-Thought Search draws significant inspiration from the formulations and methodologies of NAS, yet adapts them to the novel domain of linguistic reasoning. Analogous to how NAS searches for an optimal sequence of layers or operations to process an image \cite{BAYMURZINA202282}, our framework searches for an optimal sequence of thinking tokens (reasoning operators) to process a complex query. We explicitly define a discrete search space of reasoning operators and employ a policy network to navigate this space, mirroring the controller-based paradigms seen in RL-based NAS \cite{zoph2017neuralarchitecturesearchreinforcement}. Furthermore, our dual-factor heuristic function, which penalizes generation length to encourage efficient reasoning, directly parallels the multi-objective optimization found in resource-aware NAS methods like MnasNet \cite{tan2019mnasnetplatformawareneuralarchitecture} or EfficientNet \cite{tan2020efficientnetrethinkingmodelscaling}. However, a critical distinction lies in the granularity and dynamism of the search. While traditional NAS typically outputs a static architecture that is fixed for all dataset instances \cite{elsken2019neuralarchitecturesearchsurvey}, our method performs a dynamic, instance-wise search where the "architecture" of the reasoning path is constructed on-the-fly conditioned on the specific input query. Additionally, unlike NAS which often requires expensive retraining of the searched architecture, our method optimizes the reasoning topology during inference time, leveraging the pre-trained capabilities of the underlying Large Language Model.
Our approach also contrasts with other automated ML techniques applied to different problem domains, such as multimodal recommendation systems \cite{zhao2025dvib, zhong2024mirror}, multimodal association evaluation \cite{liu2025associam}, dialogue systems \cite{zhong2022cem}, and continual learning in NLP \cite{chen2025prototype}, which focus on optimizing specific application pipelines.

\section{Characterization of Thinking Tokens}
\label{appendix:thinking_tokens}
\subsection{Experimental Setup}
To explore the relationship between thinking tokens and think patterns, we analyze traces from two Small Reasoning Models: DeepSeek-R1-Distill-Qwen-1.5B and DeepSeek-R1-Distill-Llama-8B. We evaluate on AMC23 (standard competition math) and AIME24 (complex reasoning chains). For each model-dataset pair, we generate full reasoning traces, segmenting them into discrete steps via the \verb|"\n\n"| delimiter. We employ DeepSeek-V3 and GPT-4o to annotate the thinking mode of each step. To ensure robustness, we utilize two prompt strategies: (1) definition-based, classifying steps against rigorous academic definitions, and (2) function-based, assessing the step's role in the problem-solving flow (see Fig. \ref{appendix:prompt1} and \ref{appendix:prompt2}). 

\subsection{Analysis of Results}
We observe a deterministic correspondence between specific initial tokens and the subsequent reasoning trajectory. The token "Wait" serves as a trigger for self-correction, initiating Reflection steps with a probability exceeding 90\%. Similarly, "Alternatively" functions as a dedicated branch indicator, leading to Divergence steps in over 95\% of cases. In contrast, "Thus" acts as a transitional operator, distributing its probability mass nearly equally between deductive Statements and conclusive Summaries. This distribution suggests that thinking tokens function not merely as syntactic connectors but as semantic control signals that modulate the generation logic. We posit that distinct initial tokens activate specific, latent thinking modes inherent to the LRM. These modes are not explicitly defined but emerge as clustered behaviors within the model's high-dimensional representation space. Our Neural Chain-of-Thought Search exploits this structure. By discretely selecting the optimal thinking token at each decision point, the algorithm effectively performs dynamic cognitive switching. This mechanism allows the system to navigate the solution space by actively engaging the most appropriate latent reasoning mode for the current context, thereby maximizing solution efficiency.

\begin{figure}[t]
  \centering
\includegraphics[width=0.98\linewidth]{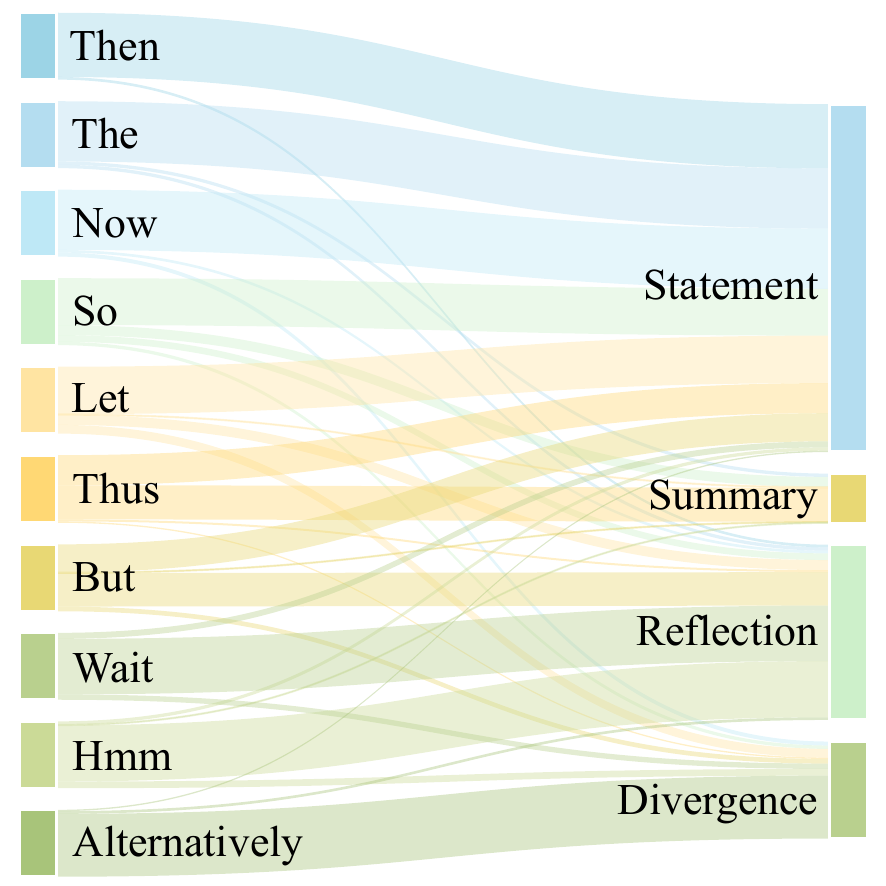}
\vspace{-5pt}
  \caption{Correlation between thinking tokens and thinking modes for DeepSeek-R1-Distill-Qwen-1.5B on the AMC23 dataset. The reasoning steps were classified by DeepSeek-V3 using the definition-based prompt strategy (Prompt 1).}
  \vspace{-15pt}
\label{fig:sankey_1}
\end{figure}

\begin{figure}[t]
  \centering
\includegraphics[width=0.98\linewidth]{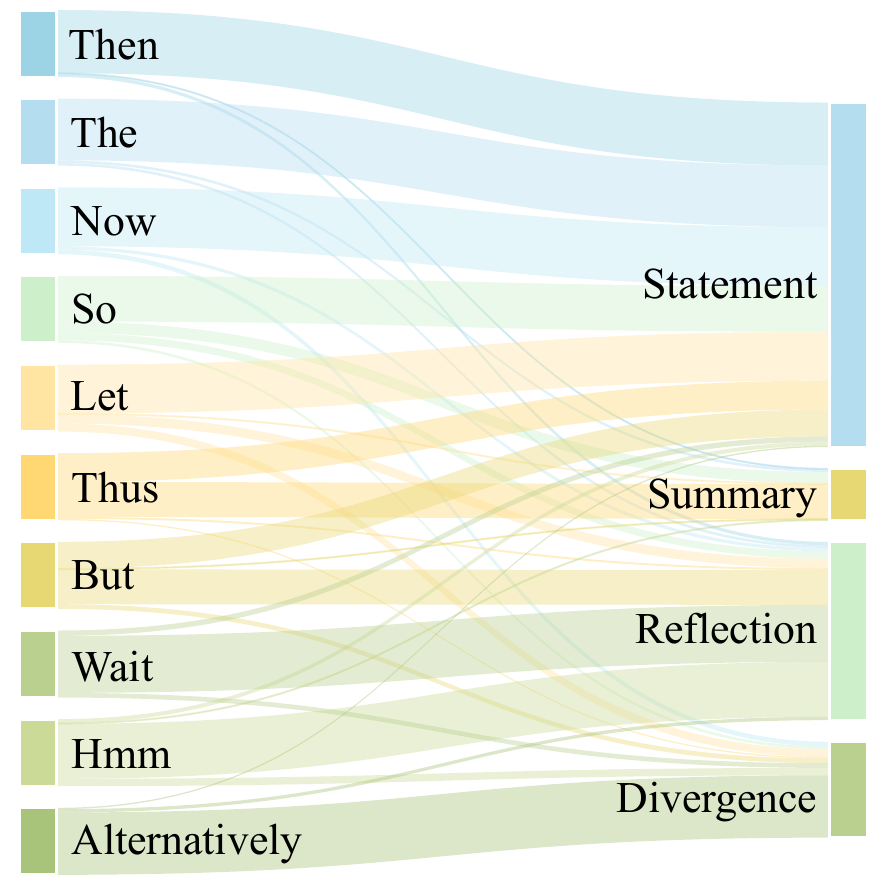}
\vspace{-5pt}
  \caption{Correlation between thinking tokens and thinking modes for DeepSeek-R1-Distill-Llama-8B on the AIME24 dataset. The reasoning steps were classified by GPT-4o using the function-based prompt strategy (Prompt 2).}
  \vspace{-15pt}
\label{fig:sankey_2}
\end{figure}

\section{Justification of Design Choices}
\label{appendix:design_choice}

\subsection{Why do we use \texttt{\textbackslash n\textbackslash n} as delimiter}
\label{appendix:delimiter}

\subsubsection{Statistical Evidence} 
Our choice of the double newline (\texttt{"\textbackslash n\textbackslash n"}) as the delimiter for reasoning steps is grounded in the empirical analysis of Large Reasoning Models' generation patterns. Recent work \cite{yang2025speculativethinkingenhancingsmallmodel} studies the distribution of tokens preceding “reasoning-supportive” keywords, which are terms that explicitly signal a shift in reasoning mode, including “Wait” (reflection), “Alternatively” (branching), and “Hmm” (hesitation). Their analysis on the MATH500 dataset reveals a strong conditional dependence: these pivot tokens are overwhelmingly preceded by the \texttt{"\textbackslash n\textbackslash n"} delimiter. As shown in Table \ref{tab:preceding_tokens}, for the token "Wait", approximately 90\% of occurrences follow a double newline. Similarly, "Alternatively" follows this delimiter pattern with a probability of over 92\%. This statistical dominance indicates that \texttt{"\textbackslash n\textbackslash n"} is not merely a syntactic formatter but a latent signal where the model naturally pauses to determine the trajectory of the subsequent thought process.

\begin{table}[t]
\centering
\begin{tabular}{llc}
\toprule
\textbf{Keyword} & \textbf{Previous Token} & \textbf{Probability} \\
\midrule
\multirow{3}{*}{"alternatively"} & \texttt{"x\textbackslash n\textbackslash n"} & $0.950$ \\
 & " " & $0.050$ \\
 & Other & $<0.001$ \\
\midrule
\multirow{3}{*}{"hmm"} & " " & $0.690$ \\
 & \texttt{"x\textbackslash n\textbackslash n"} & $0.297$ \\
 & Other & $0.013$ \\
\midrule
\multirow{3}{*}{"wait"} & \texttt{"x\textbackslash n\textbackslash n"} & $0.808$ \\
 & " " & $0.182$ \\
 & Other & $0.100$ \\
\bottomrule
\end{tabular}
\caption{Proportion of top preceding tokens for reasoning-supportive words in Deepseek-Distilled Qwen-2.5-32B on MATH500. Data adapted from previous work \cite{yang2025speculativethinkingenhancingsmallmodel}. The results show that the double newline delimiter dominates the distribution. 
\texttt{"x\textbackslash n\textbackslash n"} denotes all tokens containing "\textbackslash n\textbackslash n", including "\textbackslash n\textbackslash n", ").\textbackslash n\textbackslash n", "?\textbackslash n\textbackslash n", " \textbackslash n\textbackslash n", "].\textbackslash n\textbackslash n", "\textbackslash n\textbackslash n", ")\textbackslash n\textbackslash n", "]\textbackslash n\textbackslash n", "?)\textbackslash n\textbackslash n".}
\vspace{-8pt}
\label{tab:preceding_tokens}
\end{table}

\subsubsection{Functional Role as Discourse Marker} 
Beyond statistical correlation, the \texttt{"\textbackslash n\textbackslash n"} token functions as a critical discourse marker in the latent space of LRMs. The segment immediately following a double newline typically determines the functional category of the next step. Analysis categorizes these post-delimiter segments into distinct modes: Affirmation (continuing the current logic), Reflection (backtracking or verifying), or Statement (deriving new formulas). For instance, when a model generates \texttt{"\textbackslash n\textbackslash n"}, it enters a "decision state" where it must implicitly choose whether to proceed or reflect. In standard autoregressive decoding, this choice is made probabilistically based on the preceding context. However, smaller models often fail at this juncture, producing repetitive "Statement" steps when a "Reflection" is required, or entering verification loops unnecessarily. 

\subsubsection{Implications for our work} 
These findings validate our formulation of the search space defined in Section \ref{sec:preliminary}. By designating \texttt{"\textbackslash n\textbackslash n"} as the decision point $d_t$, we align our search intervention with the model's intrinsic cognitive structure. Rather than imposing arbitrary boundaries, we intervene exactly at the moment the model naturally pauses to select a thinking mode. Our method effectively externalizes this implicit decision-making process. By injecting explicit operators (e.g., forcing a "Wait" or "So") at these precise structural breaks, we can actively steer the model out of suboptimal paths (such as the Excessive Reflection identified in smaller models) and toward the superior paths that optimize both accuracy and efficiency.

\subsection{How do we choose the operator set $\mathcal{O}$}
\label{appendix:operator_set}
We determine the composition of the operator set $\mathcal{O}$ through a statistical analysis of the vocabulary distribution in the training corpus. Specifically, we calculate the frequency of tokens that immediately follow the step delimiter \verb|"\n\n"|. By selecting the tokens that most frequently initiate a new reasoning step, we ensure that our constructed search space aligns with the model's natural generation patterns and covers the most probable reasoning transitions.

\subsubsection{Operator Set for Neural CoT Search}
In our proposed method, we employ a comprehensive operator set $\mathcal{O} = \{$``The'', ``Thus'', ``Therefore'', ``So'', ``Then'', ``Let'', ``Wait'', ``Alternatively'', ``Now'', ``I'', ``First'', ``Option'', ``**'', ``-'', ``\textbackslash['', ``\textbackslash''$\}$. We selected these tokens because they are not only statistically frequent but also serve distinct and necessary functional roles in structuring the reasoning process. Beyond the standard thinking tokens that guide logical flow (e.g., ``Thus'', ``Wait''), we explicitly include functional markers. For instance, ``Option'' is crucial for analyzing specific choices in multiple-choice questions; ``**'' and ``-'' are widely used for emphasis and enumeration to organize complex arguments; and ``\textbackslash['' and ``\textbackslash'' are essential for initiating mathematical derivation blocks. Each of these tokens represents a specific mode of operation that the model frequently utilizes to construct valid reasoning steps.

\subsubsection{Operator Set for Random Search}
In contrast, for our random search experiments, we utilize a restricted subset: $\mathcal{O}_{\text{random}} = \{$``The'', ``Thus'', ``Therefore'', ``So'', ``Then'', ``Let'', ``Wait'', ``Alternatively''$\}$. The rationale for this difference lies in the inherent limitation of random sampling. Unlike our learned policy, a random search agent lacks the semantic understanding to apply context-dependent formatting tokens correctly. Randomly selecting structural tokens such as ``First'', ``Option'', ``**'', ``-'', or ``\textbackslash['' without appropriate context often leads to syntactically broken or incoherent generations (e.g., opening a math block with ``\textbackslash['' when no calculation is needed, or starting a list with ``-'' in the middle of a sentence). Therefore, we limit the random search space to purely connective thinking tokens to ensure that the sampled paths remain semantically plausible.

\subsubsection{Impact of Operator Set}
The selection of the operator set involves a fundamental trade-off between search potential and computational cost. Expanding the operator set to encompass a broader vocabulary increases the coverage of the search space, which theoretically allows for the discovery of even higher-quality reasoning paths. 
Specifically, while our current set is primarily optimized for English STEM reasoning, the framework allows for straightforward extension to multilingual domains by incorporating thinking tokens from other languages. Furthermore, the set can be recalibrated to support creative tasks by adding tokens that guide narrative planning or brainstorming.
However, a larger set also increases the branching factor at each decision point, which raises both the training complexity of the policy network and the computational overhead during inference. Our empirical observations indicate that the current selection strikes an effective balance, enabling the discovery of superior reasoning paths while keeping the search process computationally efficient.

\begin{table*}[t]
\centering
\small
\begin{adjustbox}{valign=c, scale=0.99}
    \begin{tabular}{ p{16cm} }
        \toprule
        % \rowcolor{myGreen}
        \textbf{Question}: ``Plants have specialized tissues at the ends of their stems. What do these tissues help the plants do? 
        
        Choices:
        A. digest food
        B. grow taller
        C. make food
        D. absorb water'' (Answer: B)  \\
        \midrule[0.5pt]
        \textcolor{blue}{\textbf{Original (2032 tokens):}}
        ``
        Okay, so I have this question about plants and their specialized tissues \ldots
        ~\colorbox{myYellow}{First}, the main types of tissues in plants are probably similar to animals 
        ~\colorbox{myBlue}{Wait}, no, the epidermis is the outer layer of the plant, \ldots
        ~\colorbox{myBlue}{But} at the ends of stems, maybe there are specific tissues \ldots
        ~\colorbox{myYellow}{Let} me think about what each tissue does. Root hair cells are \ldots
        ~\colorbox{myYellow}{Oh}, right! The ends of stems are called leaves. \ldots
        ~\colorbox{myBlue}{Wait}, the question is about specialized tissues at the ends of their stems. \ldots
        ~\colorbox{myBlue}{Wait}, maybe it's the xylem or phloem at the tip?  \ldots
        ~\colorbox{myBlue}{But} wait, the question says ""at the ends of their stems."" Leaves are at the ends of branches, not just stems \ldots
        ~\colorbox{myBlue}{But} I'm a bit confused because leaves are the main site for photosynthesis, which involves making food. 
        ~\colorbox{myYellow}{So}, maybe the answer is C. 
        ~\colorbox{myBlue}{Wait}, the question says ""the ends of their stems."" \ldots
        ~\colorbox{myGreen}{Alternatively}, maybe the question is referring to the terminals of the stem, which are where leaves grow.
        ~\colorbox{myYellow}{So}, the cells there might be specialized for photosynthesis. 
        ~\colorbox{myYellow}{So}, if the cells are specialized for making food, then the answer is C.
        ~\colorbox{myBlue}{But} I'm not entirely sure. Let me think again. \ldots
        ~\colorbox{myYellow}{So}, the ends of stems \ldots
        ~\colorbox{myYellow}{Then}, the mesophyll is where photosynthesis occurs \ldots
        ~\colorbox{myYellow}{So}, the tissues themselves are the epidermis and mesophyll, \ldots
        ~\colorbox{myBlue}{Wait}, maybe I'm overcomplicating it. \ldots
        ~\colorbox{myYellow}{So}, the question is asking what these tissues help the plant do \ldots
        ~\colorbox{myBlue}{Wait}, perhaps the answer is both D and C.
        ~\colorbox{myYellow}{Looking} at the options: A. digest food \ldots
        ~\colorbox{myYellow}{So}, if the specialized tissues at the ends of stems are leaves \ldots
        ~\colorbox{myGreen}{Alternatively}, maybe it's the xylem or phloem at the ends of stems. \ldots
        ~\colorbox{myBlue}{But} I'm still a bit confused because I know that the epidermis is specialized for absorbing water \ldots
        ~\colorbox{myYellow}{I} think I need to decide. If the question is about the specialized tissues at the ends of stems, \ldots
        ~\colorbox{myBlue}{But} if it's about the epidermis, then D.
        ~\colorbox{myYellow}{I} think the correct answer is D, absorb water, because \ldots
        ~\colorbox{myYellow}{I} think I'll go with D, absorb water.
        ~**Final Answer** \boxed{D}
        ''\\
        \midrule[0.5pt]
        \textcolor{red}{\textbf{NCoTS (433 tokens):}} 
        ``
        Okay, so I have this question about plants and their specialized tissues. \ldots
        ~\colorbox{myYellow}{First}, I remember that plants have different kinds of tissues. \ldots
        ~\colorbox{myBlue}{Wait}, what do those tissues do? I recall that in the stems, the ends are responsible for \ldots
        ~\colorbox{myYellow}{Let} me go through the options again to make sure. 
        ~\colorbox{myYellow}{Option} A is digest food. That's more about root hairs \ldots
        ~\colorbox{myYellow}{Option} B is grow taller. As I thought earlier \ldots
        ~\colorbox{myYellow}{Option} C is make food. Making food is done by leaves \ldots
        ~\colorbox{myYellow}{Option} D is absorb water. I think water absorption is more related to the roots \ldots
        ~\colorbox{myYellow}{So}, putting it all together, the correct answer should be B \ldots
        ~~**Final Answer** \boxed{B}
        ''\\
        \bottomrule
    \end{tabular}
\end{adjustbox}
\caption{\small\setlength{\baselineskip}{1.25\baselineskip}\looseness=-1 Comparison of reasoning paths between original DeepSeek-R1-Distill-Qwen-7B (top) and our NCoTS (bottom). 
\colorbox{myBlue}{Blue} boxes: reflection/verification steps. 
\colorbox{myGreen}{Green} boxes: divergence steps.
\colorbox{myYellow}{Yellow} boxes: statement/summarization steps.
The original model drifts into irrelevant reasoning branches, resulting in an incorrect answer.
NCoTS enables the model to reach the correct conclusion with higher efficiency.}
\label{tab:case1}
% \vspace{-0.5em}
\end{table*}

\section{Case Study}
We present two qualitative comparisons between the original generation and our proposed NCoTS. As detailed in Table \ref{tab:case1} and Table \ref{tab:case2}, these cases illustrate how current reasoning models frequently lack foresight and consequently fail to navigate the solution space efficiently.

In the commonsense science query shown in Table \ref{tab:case1}, the original model experienced severe reasoning drift. Lacking a high level plan, it conflates concepts like ends of stems with leaves and wanders into irrelevant topics such as water absorption, which ultimately leads to an incorrect conclusion
In contrast, NCoTS mitigates this issue by actively searching for optimal thinking mode at decision points. This mechanism allows the model to prune these inefficient branches early, effectively steering the path toward the correct solution with significantly reduced token usage.
Similarly, the mathematical reasoning task in Table \ref{tab:case2} highlights the inefficiency of myopic next token prediction. 
While the original model eventually finds the solution, it suffers from a lack of confidence, evidenced by getting stuck in redundant verification loops and exploring complex but unnecessary branches. 
NCoTS eliminates this overhead by prioritizing decisive operators that advance the solution state. Our approach performs one succinct verification step, reflecting a more confident and efficient reasoning process.

These case studies form a closed loop with our initial motivation. They confirm that the inefficiency of current reasoning models stems from the failure to foresee the appropriate reasoning direction at decision points. By treating reasoning as a dynamic search for the optimal thinking mode, NCoTS actively selects the suitable reasoning direction and effectively prunes redundant branches. This yields a superior reasoning path that is both accurate and concise, avoiding reasoning traps such as getting stuck in redundant verification loops or excessive exploration.

\begin{table*}[t]
\centering
\small
\begin{adjustbox}{valign=c, scale=0.99}
    \begin{tabular}{ p{16cm} }
        \toprule
        % \rowcolor{myGreen}
        \textbf{Question}: There is a unique sequence of integers $a_1, a_2, \cdots a_{2023}$ such that\[\tan2023x = \frac{a_1 \tan x + a_3 \tan^3 x + a_5 \tan^5 x + \cdots + a_{2023} \tan^{2023} x}{1 + a_2 \tan^2 x + a_4 \tan^4 x \cdots + a_{2022} \tan^{2022} x}\]whenever $\tan 2023x$ is defined. What is $a_{2023}?$ (Answer: -1)  \\
        \midrule[0.5pt]
        \textcolor{blue}{\textbf{Original (4416 tokens):}}
        ``Okay, so I have this problem here where I need to find the value of \( a_{2023} \) in a unique sequence of integers. \ldots
        ~\colorbox{myBlue}{Hmm}, okay. I remember that multiple-angle identities for tangent can be expressed using \ldots
        ~\colorbox{myYellow}{Given} that, I think the coefficients \( a_1, a_3, \ldots, a_{2023} \) are related to \ldots
        ~\colorbox{myGreen}{Alternatively}, I remember that there is a formula involving Bernoulli numbers or some recursive relation for the coefficients in the multiple-angle expansion of tangent. \ldots
        ~\colorbox{myGreen}{Wait}, another thought: the tangent function can be expressed using complex numbers. \ldots
        ~\colorbox{myGreen}{Alternatively}, perhaps using the expansion in terms of Bernoulli polynomials or some generating function? \ldots
        ~\colorbox{myBlue}{Wait}, maybe I can use the identity for \( \tan(nx) \). \ldots
        ~\colorbox{myYellow}{Yes}, for example, for \( \tan(3x) \), as I mentioned earlier, it's \( \frac{3 \tan x - \tan^3 x}{1 - 3 \tan^2 x} \). \ldots
        ~\colorbox{myBlue}{Wait}, looking at \( \tan(3x) \), the numerator is \( 3 \tan x - \tan^3 x \), which is similar to the expansion of \( \sin(3x)/\cos(3x) \), \ldots
        ~\colorbox{myBlue}{But} perhaps for the general case, the expansion of \( \tan(nx) \) as a rational function \ldots
        ~\colorbox{myGreen}{Wait}, another approach: perhaps using the formula for \( \tan(A + B) \). \ldots
        ~\colorbox{myBlue}{But} 2023 is a large number, so recursively applying this would be tedious. \ldots
        ~\colorbox{myYellow}{After} a quick recall, I remember \ldots
        ~\colorbox{myYellow}{Moreover}, the leading term in the numerator is \ldots
        ~\colorbox{myBlue}{Wait}, let's test this with small n. \ldots
        ~\colorbox{myYellow}{So}, in general, it seems that for \( \tan(nx) \) \ldots
        ~\colorbox{myBlue}{Wait}, but in the case of n=3 \ldots
        ~\colorbox{myYellow}{Therefore}, for n=2023, the leading term in the numerator would be \ldots
        ~\colorbox{myBlue}{Wait}, but in the problem statement \ldots
        ~\colorbox{myBlue}{But} hold on, let me check with n=3 again. \ldots
        ~\colorbox{myBlue}{Wait}, that's a problem. \ldots
        ~Looking at n=3 \ldots
        ~\colorbox{myYellow}{So}, for n=3, the leading coefficient is -1 \ldots
        ~\colorbox{myBlue}{Wait}, for n=3: \( (-1)^{(3-1)/2} = (-1)^1 = -1 \) \ldots
        ~\colorbox{myBlue}{But} in n=3, the leading coefficient is -1, which matches \( (-1)^1 = -1 \). \ldots
        ~\colorbox{myYellow}{So}, for n=2023, the leading coefficient is -1, so \( a_{2023} = -2023 \). \ldots
        ~\colorbox{myBlue}{Wait}, no, in n=3, the leading term is \( -\tan^3 x \), so the coefficient is -1. \ldots
        ~\colorbox{myYellow}{Therefore}, for n=2023 \ldots
        ~\colorbox{myBlue}{But} in the numerator, for n=3 \ldots
        ~\colorbox{myYellow}{So}, in general, for the numerator of \ldots
        ~\colorbox{myBlue}{Wait}, hold on. In n=3, the leading term is \ldots
        ~\colorbox{myBlue}{Wait}, but in the problem statement \ldots
        ~\colorbox{myBlue}{But} hold on, in n=3 \ldots
        ~\colorbox{myYellow}{{So}}, the coefficient of \( \tan^n x \) is \ldots
        ~\colorbox{myYellow}{Therefore}, for n=2023, the coefficient \( a_{2023} \) is \( (-1)^{(2023 - 1)/2} = (-1)^{1011} = -1 \). \ldots
        ~\colorbox{myBlue}{But} hold on a second, in n=3, the leading term is \( -\tan^3 x \) \ldots
        ~\colorbox{myYellow}{Yes}, \( \tan(3x) = \frac{3 \tan x - \tan^3 x}{1 - 3 \tan^2 x} \). \ldots 
        ~\colorbox{myYellow}{Similarly}, for n=5, the numerator is \( 5 \tan x - 10 \tan^3 x + \tan^5 x \) \ldots
        ~\colorbox{myYellow}{So}, the leading term is \( -\tan^3 x \). So, the coefficient is -1. \ldots
        ~\colorbox{myYellow}{So}, in general, for each odd n \ldots
        ~\colorbox{myBlue}{Wait}, but in n=3, the leading coefficient is -1 \ldots
        ~\colorbox{myYellow}{Therefore}, \( a_{2023} = (-1)^{(2023 - 1)/2} = (-1)^{1011} = -1 \). \ldots
        ~\colorbox{myBlue}{But} wait, another thought \ldots
        ~\colorbox{myBlue}{But} let's think about how the expansion is constructed \ldots
        ~\colorbox{myYellow}{So}, based on the pattern observed in smaller cases \ldots
        ~**Final Answer** \boxed{-1}''\\
        \midrule[0.5pt]
        \textcolor{red}{\textbf{NCoTS (1771 tokens):}} 
        ``
        Okay, so I have this problem here about a unique sequence of integers \( a_1, a_2, \ldots, a_{2023} \) \ldots
        ~I need to find the value of \( a_{2023} \).
        ~\colorbox{myYellow}{First}, I recall that there's a formula for the tangent of multiple angles. \ldots
        ~\colorbox{myYellow}{So}, for \( n = 2023 \), the numerator will be a polynomial of degree 2023 in \( \tan x \), \ldots
        ~\colorbox{myYellow}{The} problem is asking for the coefficient \( a_{2023} \) of the highest degree term \ldots
        ~\colorbox{myYellow}{Now}, how do I find this coefficient?
        ~\colorbox{myYellow}{I} remember that the tangent multiple-angle formula can be derived using the addition formula for tangent. The addition formula is:
        ~$\tan(A + B) = \frac{\tan A + \tan B}{1 - \tan A \tan B}$
        ~\colorbox{myYellow}{So}, if we apply this repeatedly, we can express \( \tan(nx) \) in terms of \( \tan x \). \ldots
        ~\colorbox{myYellow}{Therefore}, in general, for \( \tan(nx) \), the numerator is:
        ~$\sum_{k=1}^{n} (-1)^{k-1} \binom{n}{2k-1} \tan^{2k-1}x$
        ~\colorbox{myYellow}{So}, the coefficient \( a_{2023} \) is \( (-1)^{1011} \binom{2023}{2023} \) \ldots
        ~\colorbox{myYellow}{Let}'s compute this: $a_{2023} = (-1)^{1011} \binom{2023}{2023}$
        ~I know that \( \binom{2023}{2023} = 1 \), so this simplifies to: \ldots
        ~\colorbox{myYellow}{Now}, \( 1011 \) is an odd number, so \( (-1)^{1011} = -1 \).
        ~\colorbox{myYellow}{Therefore}, \( a_{2023} = -1 \).
        ~\colorbox{myBlue}{Wait}, let me just verify if this makes sense. \ldots
        ~\colorbox{myYellow}{So}, the numerator is \( 3 \tan x - \tan^3 x \) \ldots
        ~\colorbox{myYellow}{So}, similarly, for \( n = 2023 \), \( a_{2023} = (-1)^{1011} \times 1 = -1 \). So, that seems consistent.
        ~I think that's the answer.
        ~\colorbox{myYellow}{The} coefficient of the highest degree term in the numerator of \( \tan 2023x \) is \( (-1)^{1011} \times 1 = -1 \).
        ~**Final Answer** \boxed{-1}
        ''\\
        \bottomrule
    \end{tabular}
\end{adjustbox}
\caption{\small\setlength{\baselineskip}{1.25\baselineskip}\looseness=-1 Comparison of reasoning paths between original DeepSeek-R1-Distill-Qwen-1.5B (top) and our NCoTS (bottom). 
\colorbox{myBlue}{Blue} boxes: reflection/verification steps. 
\colorbox{myGreen}{Green} boxes: divergence steps.
\colorbox{myYellow}{Yellow} boxes: statement/summarization steps.
This case demonstrates that the original model struggles with a lack of foresight, leading to excessive exploration and redundant verification.
In contrast, NCoTS delivers a coherent derivation that solves the problem using fewer than 50\% of the tokens.}
\label{tab:case2}
% \vspace{-0.5em}
\end{table*}

\end{document}